%% file: iclr2023_workshop.tex
\documentclass{article} 
\usepackage{iclr2023_workshop,times}

\input{math_commands.tex}

\usepackage[colorlinks=true,citecolor=brown,urlcolor=gray]{hyperref}
\usepackage{url}
\usepackage{graphicx}
\usepackage{comment}
\usepackage{booktabs}
\usepackage{makecell}
\usepackage[linesnumbered,ruled]{algorithm2e}
\usepackage{amsmath}
\usepackage{xcolor}
\usepackage{adjustbox}

\newcommand{\mat}[1]{\mathbf{#1}}
\newcommand{\zm}[1]{{\color{black!0!blue} #1}}

\title{GenPhys: From Physical Processes to Generative Models}

\author{Ziming Liu, Di Luo, Yilun Xu, Tommi Jaakkola, Max Tegmark \\
Massachusetts Institute of Technology \\
\texttt{\{zmliu,diluo,ylxu,jaakkola,tegmark\}@mit.edu}}

\iclrfinalcopy 
\begin{document}

\maketitle

\begin{abstract}
Since diffusion models (DM) 
and the more recent Poisson flow generative models (PFGM) are inspired by physical processes, it is reasonable to ask: {\bf Can physical processes offer additional new generative models?} We show that the answer is {\bf Yes}.  We introduce a general family, {\bf Gen}erative Models from {\bf Phys}ical Processes ({\bf GenPhys}), where we translate partial differential equations (PDEs) describing physical processes to generative models. We show that generative models can be constructed from \textit{s-generative} PDEs (\textit{s} for smooth).
GenPhys subsume the two existing generative models (DM and PFGM) and even give rise to new families of  generative models, e.g., ``Yukawa Generative Models" inspired from weak interactions. On the other hand, some physical processes by default do not belong to the GenPhys family, e.g., the wave equation and the Schr\"{o}dinger equation, but could be made into the GenPhys family with some modifications. Our goal with GenPhys is to explore and expand the design space of generative models.
\end{abstract}

\section{Introduction}

\begin{figure}[h]
    \centering
    \includegraphics[width=1\linewidth]{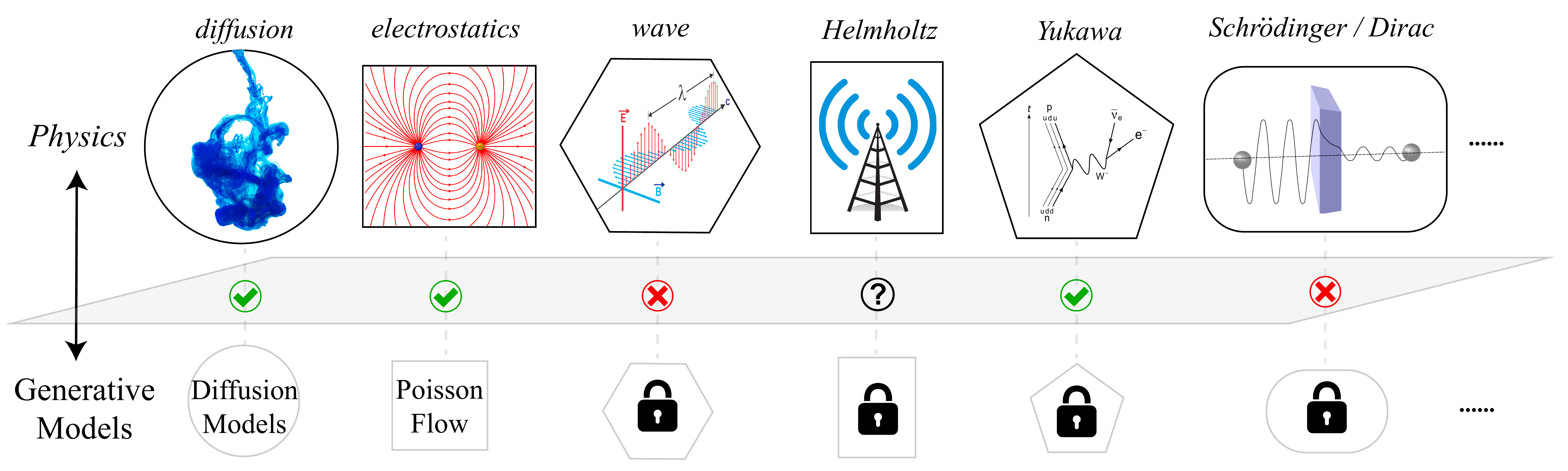}
    \caption{Duality between physics and generative models. So far only diffusion models and Poisson flow models are discovered in the literature. Can we unlock more?}
    \label{fig:duality}
\end{figure}

Recently, we have witnessed the success of physics-inspired deep generative models, such as diffusion models (DM)~\citep{sohl2015deep, ho2020denoising,song2020score,karras2022elucidating} based on thermodynamics and Poisson flow genertive models (PFGM)~\citep{xu2022poisson, Xu2023PFGMUT} derived from electrostatics. The idea of diffusion models is to reverse the process of ink diffusing in water, while PFGM view data points as charged particles and let them move in electric fields. Illustrated in Figure~\ref{fig:duality}, there seems to exist a \textit{duality} between physics and generative models, i.e., a physical phenomenon can give rise to a generative model, and vice versa. Does such duality really exist? We will show that the answer is {\bf Yes}, albeit with some restrictions on the physical processes. This duality raises the possibility of augmenting the design space of generative models with nearly no effort, simply by by leveraging the underlying dynamics of diverse physical structures, including molecules, stars, galaxies, planets, and even human beings.

The connection between physics and generative models can be quite deep. Our Universe is arguably a generative model~\citep{tegmark1996does, lin2017does}: Starting from the wave function of our early Universe, which was a simple multivariate Gaussian corresponding to spatially uniform fields with small quantum fluctuations, our Universe evolves to ``generate" ever richer 
and more complex phenomenon. However, the dynamics that drives the evolution is described by (simple and elegant) partial differential equations (PDEs). The same applies to generative models which leverage continuous physical processes: although the whole transformation from latent to data distribution can be quite complicated, the movement at each step is simple, ready to be learned by deep neural networks.

This work focuses on generative models inspired by continuous physical processes, very much sharing the flavor of DM and PFGM, but seeks a more unified framework. Since continuous physical processes are described by PDEs, we will use these two terms interchangeably. We propose a framework that 
can convert physical PDEs to generative models, termed {\bf Gen}erative Models from {\bf Phys}ical Processes ({\bf GenPhys}). Specifically, for a PDE $p$~\footnote{All the PDEs discussed in this paper are equipped with free boundary conditions.}, we denote the corresponding generative model $p$-GenPhys. For example, diffusion models and Poisson flow generative models leverage the diffusion equation and the Poisson equation, respectively, so they are called diffusion-GenPhys and Poisson-GenPhys under the GenPhys framework. We will show that $p$-GenPhys is a generative model if the PDE $p$ is \textit{s-generative} (\textit{s} for smooth), meaning that these two conditions are met: 

\begin{itemize}
  \item[\bf] {\bf (C1)} $p$ is equivalent to a density flow;
  \item[\bf] {\bf (C2)} The solution of $p$ becomes ``smoother" over time.
\end{itemize}
Although {\bf (C2)} is handwavy, it can be made rigorous with dispersion relations (see Section \ref{sec:dispersive-relation}), the main idea being that ``smoothing" means high-frequency modes decay faster than low-frequency modes. With these two conditions, we can thus categorize $p$ into three classes - s-generative, conditionally s-generative (depending on some coefficients in PDE), or not s-generative: 
\begin{itemize}
  \item[\bf] {\bf (1)} $p$ is s-generative. Examples: $p=$ diffusion, Poisson, Yukawa (screened Poisson), biharmonic, fractal diffusion, higher-order diffusion.
  \item[\bf] {\bf (2)} $p$ is  conditionally s-generative. Examples: $p=$ dissipative wave, Helmholtz.
  \item[\bf] {\bf (3)} $p$ is not s-generative. Examples: $p=$ ideal wave, Schr\"{o}dinger, Dirac.
\end{itemize}

The rest of the paper is organized as follows: Section \ref{sec:condition} introduces the GenPhys framework that converts physical processes to generative models. Section \ref{sec:processes} goes through common physical processes, demonstraing the GenPhys framework on these PDEs. Section \ref{sec:dispersive-relation} proposes to use a rigorous criterion, dispersion relations, to determine whether a PDE is smooth. Section \ref{sec:related_works} reviews related works, followed by conclusions and discussions in Section \ref{sec:conclusion}.

\section{Generative Models from Physical Processes (GenPhys)}\label{sec:condition}

This section reveals a connection between continuous physical processes and generative models. The key is to match their associated PDEs: each physical process is described by a PDE, while each generative model is associated with a density flow (which is also a PDE). To convert a physical process to a generative model, we need the following conversion steps:

\begin{equation}
\tiny
    \boxed{{\rm Physical\ process}}  \xrightarrow[{\rm Sec}\ \ref{subsec:physical-pde}]{(a)}\boxed{{\rm Physical\ PDE\ (Eq.\ref{eq:physical-PDE})}} 
     \xrightarrow[{\rm Sec}\ \ref{subsec:converting}]{(b)}\boxed{{\rm Density\ flow\ (Eq.\ref{eq:density-flow})} }\xrightarrow[{\rm Sec}\ \ref{subsec:density-flow}]{(c)}\boxed{{\rm Generative\ model}}
\vspace{5pt}
\end{equation}
The step (a) and (c), discussed in Section \ref{subsec:physical-pde} and \ref{subsec:density-flow}, are relatively  straightforward. In short, the step (a) holds since partial differential equations (mathematical objects) are just abstractions of physical processes (physical entities). In fact, one can start from any PDE regardless of its meaning, although we focus on PDEs that have physical meaning. The step (c) is straightforward, since we focus on the mathematical structures of generative models, ignoring their implementation details. Henceforth, we will not differentiate between ``physical process" and ``physical PDE", nor between ``density flow" and``generative model". The only challenge left is (b), i.e., converting a physical PDE to a density flow, which will be discussed in Section \ref{subsec:converting}. 

\begin{algorithm}[t]
    \SetKwInOut{Input}{Input}
    \SetKwInOut{Output}{Output}

    \Input{partial differential equation $\hat{L}\phi(\mat{x},t)=0$, data distribution $p_{\rm data}(\mat{x})$}
    \Output{generated samples $\mat{x}$}
    (1) Rewrite $\hat{L}\phi(\mat{x},t)$ in the form $\frac{\partial p(\mat{x},t)}{\partial t}+\nabla\cdot[p(\mat{x},t)\mat{v}(\mat{x},t)]-R(\mat{x},t)$  such that $p=p(\phi,\phi_t,\nabla\phi,\cdots)$, $\mat{v}=\mat{v}(\phi,\phi_t,\nabla\phi,\cdots)$, $R=R(\phi,\phi_t,\nabla\phi,\cdots)$  \;
    (2) Solve $\hat{L}\phi(\mat{x},t)=p_{\rm data}(\mat{x})\delta(t)$. If $\hat{L}$ is linear, we can express $\phi(\mat{x},t)$ in terms of the Green's function $G(\mat{x},t;\mat{x}')$: $\phi(\mat{x},t)=\int G(\mat{x},t;\mat{x}')p_{\rm data}(\mat{x}')d^N\mat{x}'$, where $\hat{L}G(\mat{x},t;\mat{x}')=\delta(\mat{x}-\mat{x}')\delta(t)$\;
    (3) Using the relations in (1) and solutions in (2) to obtain $p(\mat{x},t)$, $\mat{v}(\mat{x},t)$, $R(\mat{x},t)$\;
    (4) Train a neural network $\mat{s}_\theta(\mat{x},t)$ to fit $\mat{v}(\mat{x},t)$ such that $\mat{s}_\theta(\mat{x},t)\approx \mat{v}(\mat{x},t)$.  Train another neural network $W_\alpha(\mat{x},t)$ to fit $R(\mat{x},t)$ such that $W_\alpha(\mat{x},t)\approx R(\mat{x},t)$\;
    (5) Draw $\mat{x}(T)\sim p(\mat{x},T)$, simulate $\frac{d\mat{x}(t)}{dt}=\mat{s}_\theta(\mat{x},t)$ from $t=T$ to $t=0$ with the branching process $W_\alpha(\mat{x},t)$. Output $\mat{x}(0)$.
    \caption{Generative models from physical processes (GenPhys)}
    \label{alg:alg}
\end{algorithm}

\subsection{Generative models as density flow}\label{subsec:density-flow}
Given i.i.d. data samples from the probability distribution $p_{\rm data}(\mat{x}), \mat{x}\in\mathbb{R}^N$, the goal of generative models is to obtain new samples from $p_{\rm data}(\mat{x})$. A continuous physical process $\frac{d\mat{x}}{dt}=\mat{v}(\mat{x}, t)$ can evolve the probability distribution $p(\mat{x},t)$ as
\begin{equation}\label{eq:continuity}
    \frac{\partial p(\mat{x},t)}{\partial t}+\nabla\cdot [p(\mat{x},t)\mat{v}(\mat{x},t)]=0,
\end{equation}
known as the \textit{probability flow equation}, or the \textit{continuity equation}. Here $p$ and $\mat{v}$ are interpreted as a probability distribution and a velocity field, respectively. The probability distribution, starting as the data distribution $p(\mat{x},0)=p_{\rm data}(\mat{x})$, evolves to a (hopefully simple) final distribution $p(\mat{x},T)$. To generate samples from $p_{\rm data}(\mat{x})$, one can first draw samples from the final distribution $p_{\rm prior}(\mat{x})\equiv p(\mat{x},T)$, and run the process $\frac{d\mat{x}}{dt}=\mat{v}(\mat{x},t)$ backward from $t=T$ to $t=0$~\footnote{Note that time is usually defined in opposite directions in physics and machine-learning applications: our Universe generates complex structures as time moves forward, whereas generative models using e.g. diffusion make things simpler over time and generate complexity by evolving backward in time.}.

Although Eq.~(\ref{eq:continuity}) assumes conservation,  we can more generally allow a non-conservative term:
\begin{equation}\label{eq:continuity-no-conservation}
    \frac{\partial p(\mat{x},t)}{\partial t} +\nabla\cdot[p(\mat{x},t)\mat{v}(\mat{x},t)] - R(\mat{x},t)=0,
\end{equation}
where $R(\mat{x},t)$ corresponds to birth/death. $R>0$ ($R<0$) means particles are born (die) in the forward process, and die (are born) in the backward process~\footnote{In the time interval $[t,t+dt]$, a forward particle at $\mat{x}$ has probability $|R|dt$ to turn into two/zero particles when $R>0$/$R<0$. }. In this case, $p(\mat{x},t)$ is a density distribution instead of probability distribution, so we call Eq.~(\ref{eq:continuity-no-conservation}) the \textit{density flow equation}. The extension of density flows has found numerous applications in machine learning, such as unbalanced optimal transport methods for modeling single-cell dynamics and domain adaptation~\citep{Mroueh2020UnbalancedSD, Fatras2021UnbalancedMO}, as well as in Bayesian inference for probabilistic modeling~\citep{Lu2019AcceleratingLS}. In practice, the density flow with birth/death dynamics can be simulated efficiently. For example, in diffusion Monte Carlo, the birth/death processes can be included as branching processes with population control on $R(\mat{x},t)$~\citep{martin_reining_ceperley_2016, lu2019accelerating}.
 
We aim to design $p(\mat{x},t)$, $\mat{v}(\mat{x},t)$ and $R(\mat{x},t)$ in Eq.~(\ref{eq:continuity-no-conservation}) such that the initial and final boundary conditions are met: (1) $p(\mat{x},0)=p_{\rm data}(\mat{x})$; (2) $p_{\rm prior}(\mat{x})\equiv p(\mat{x},T)$ is asymptotically independent of $p_{\rm data}(\mat{x})$ as $T\to\infty$. Such a design process is highly non-trivial~\citep{lipman2022flow}, mostly due to the complicated boundary conditions and lack of analytical solutions in general. We will show that physics can inspire and thus facilitate the design process: Firstly, the boundary conditions can be nicely interpreted in physics and boil down to the dispersion relation, an important concept in physics (see Section \ref{sec:dispersive-relation}).
Secondly, many physical processes admit analytical solutions. For reasons that will soon become clear, it is convenient to include the initial condition as a source term in the RHS of Eq.~(\ref{eq:continuity-no-conservation}), and define the LHS as a differential operator $\hat{M}$ acting on $p(\mat{x},t)$, $\mat{v}(\mat{x},t)$ and $R(\mat{x},t)$:

\begin{equation}\label{eq:density-flow}
\boxed{
    \hat{M}(p,\mat{v},R)\equiv \frac{\partial p(\mat{x},t)}{\partial t} +\nabla\cdot[p(\mat{x},t)\mat{v}(\mat{x},t)] - R(\mat{x},t)=p_{\rm data}(\mat{x})\delta(t)
    \quad ({\rm Density\ Flow})
    }
\end{equation}

\subsection{Physical processes and physical PDEs}\label{subsec:physical-pde} 

Continuous physical processes are described by partial differential equations 
\begin{equation}\label{eq:physical-PDE}
\boxed{
 \hat{L}\phi\equiv F(\phi,\phi_t,\phi_{tt},\nabla\phi,\nabla^2\phi,...)=f(\mat{x},t)\quad {(\rm Physical\ PDE)}
 }
\end{equation}
where $\phi(\mat{x},t)$ is a scalar function defined on $\mathbb{R}^N\times \mathbb{R}^+$, $\hat{L}$ is a differential operator acting on $\phi(\mat{x},t)$, $f(\mat{x},t)$ is the source term, and subscripts stand for partial derivatives, e.g., $\phi_t\equiv\frac{\partial\phi}{\partial t}, \phi_{tt}\equiv\frac{\partial^2\phi}{\partial t^2}$. For simplicity, we will mostly discuss linear PDEs which are also symmetric both in space and time~\footnote{This means that $\hat{L}$ remains unchanged under (1) translations in time $t\to t+\delta t$, (2) translations in space $\mat{x}\to\mat{x}+\mat{\delta x}$, and (3) rotations in space $\mat{x}\to \mat{R}\mat{x}$.}, i.e., where $F$ does not depend explicitly on $\mat{x}$ or $t$. The linearity and symmetries usually make $\phi(\mat{x},t)$ analytically solvable and these solutions are available in many mathematical physics textbooks~\citep{kirkwood2018mathematical, guenther1996partial}.

For linear PDEs, the solution $\phi(\mat{x},t)$ can be expressed as a convolution of the Green's function $G(\mat{x},t;\mat{x}',t')$ with the source term $f(\mat{x},t)$, i.e., $\phi(\mat{x},t)=\int G(\mat{x},t;\mat{x}',t')f(\mat{x}',t')d^N\mat{x}'dt$~\zm{~\citep{courant2008methods}}. The Green's function $G(\mat{x},t;\mat{x}',t')$ is defined as the solution of $\hat{L}G(\mat{x},t;\mat{x}',t')=\delta(\mat{x}-\mat{x}')\delta(t-t')$ with $G(\mat{x},t;\mat{x}',t')=0$ when $t<t'$.

\subsection{Converting Physical PDEs to Density flows}\label{subsec:converting}
Recall that our goal is, given $p_{\rm data}(\mat{x})$, to design $(p,\mat{v},R)$ such that the density flows equation~(\ref{eq:density-flow}) holds. Since physical PDEs (Eq.~\ref{eq:physical-PDE}) are well studied and have nice properties, it is really convenient if we can transfer the solutions of physical PDEs to those of density flows. A hope is that density flows and physical PDEs are actually the same equation, if both LHS and RHS match. The match of RHS is easy, by simply setting $f(\mat{x},t)=p_{\rm data}(\mat{x})\delta(t)$. The match of the LHS is non-trivial, requiring that $(p,\mat{v},R)$ depend on $\phi$ in a clever way such that setting $p=p(\phi,\phi_t,\nabla\phi,\cdots)$, $\mat{v}=\mat{v}(\phi,\phi_t,\nabla\phi,\cdots)$ and $R=R(\phi,\phi_t,\nabla\phi,\cdots)$ in Eq.~(\ref{eq:density-flow}) gives $F(\phi,\phi_t,\cdots)$ in Eq.~(\ref{eq:physical-PDE}). To ensure that $(p,\mat{v},R)$ defines a density flow, we should check 

\begin{itemize}
  \item[\bf (C1)] \textit{Well-behaved density flow}: $p(\mat{x},t)$ is a density distribution, i.e., $p(\mat{x},t)\geq 0$. In addition, $(p, \mat{v}, R)$ should be well-behaved (e.g., cannot be discontinuous or have singularities etc.).
\end{itemize}

However, {\bf (C1)} is not enough for generative models. In practice, we want generative models to have a prior distribution $p_{\rm prior}(\mat{x})\equiv p(\mat{x},T)$ that is independent of $p(\mat{x},0)\equiv p_{\rm data}(\mat{x})$. Intuitively, this requires that $p(\mat{x},t)$ becomes ``smoother" as $t$ increases such that the initial details are ``hidden" or ``blurred". We can informally formulate this condition as:

\begin{itemize}\label{list:conditions}
  \item[\bf (C2)] \textit{Smooth PDEs}:  As $T\to\infty$, the final distribution $p(\mat{x},T)$ becomes asymptotically independent of the initial distribution $p(\mat{x},t=0)=p_{\rm data}(\mat{x})$.
\end{itemize}
If a PDE satisfies both {\bf (C1)} and {\bf (C2)}, we call it \textit{s-generative}, where \textit{s} stands for smooth.

{\bf How to decide if a PDE is s-generative?} To check {\bf (C1)}, we need to match the physical PDE and the density flow. Although the matching process is case-dependent, it is usually straight-forward with a little bit of construction. Below we will demonstrate it on two known cases: the diffusion equation and the Poisson equation. To check {\bf (C2)}, physical insights are usually sufficient to determine whether the PDE is smooth or not. In Section \ref{sec:processes}, we will present examples of smooth and non-smooth PDEs, where smoothness is clear from illustrations of Green's functions. We defer rigorous discussion of {\bf (C2)} to Section \ref{sec:dispersive-relation}, where we show that {\bf (C2)} is equivalent to a constraint on the dispersion relations of PDEs~\footnote{The equvalence is true for linear PDEs, the focus of this paper. For non-linear PDEs, the condition becomes more involved.}. Putting everything together, Alg.~\ref{alg:alg} summarizes the ``algorithm" that leverages a physical PDE to generate samples.

{\bf Example 1: Diffusion models} We aim to convert the diffusion equation $\phi_t-\nabla^2\phi=p_{\rm data}(\mat{x})\delta(t)$ to a density flow $\frac{\partial p}{\partial t}+\nabla\cdot(p\mat{v})-R=p_{\rm data}(\mat{x})\delta(t)$:
\begin{equation}
    \phi_t - \nabla^2\phi = \frac{\partial\textcolor{red}{\phi}}{\partial t} + \nabla\cdot(\textcolor{red}{\phi}(\textcolor{blue}{-\nabla{\rm log}\phi})) - \textcolor{green}{0}=0\quad \Leftrightarrow\quad  \frac{\partial \textcolor{red}{p}}{\partial t}+\nabla\cdot(\textcolor{red}{p}\textcolor{blue}{\mat{v}})-\textcolor{green}{R}=0
\end{equation}
Comparing the two sides gives:
\begin{equation}\label{eq:diffusion-match}
    p = \phi,\quad \mat{v} = -\nabla{\rm log}\phi,\quad R=0
\end{equation}
The solution $\phi$ of the diffusion equation is:
\begin{equation}\label{eq:diffusion-phi-solution}
    \phi(\mat{x},t) = \int G(\mat{x},t;\mat{x}')p_{\rm data}(\mat{x}')d^N\mat{x}',\quad G(\mat{x},t;\mat{x}')=\frac{1}{(2\pi t)^{N/2}}{\rm exp}(-\frac{|\mat{x}-\mat{x}'|^2}{2t})
\end{equation}
Combining Eq.~(\ref{eq:diffusion-match}) and Eq.~(\ref{eq:diffusion-phi-solution}) gives
\begin{equation}
    \begin{aligned}
        p(\mat{x}, t) & = \phi(\mat{x},t) = \int G(\mat{x},t;\mat{x}')p_{\rm data}(\mat{x}')d^N\mat{x}'=\frac{1}{(2\pi t)^{N/2}} \int p_{\rm data}(\mat{x}'){\rm exp}(-\frac{|\mat{x}-\mat{x}'|^2}{2t})d^N\mat{x}' \\
        \mat{v}(\mat{x},t) & = -\nabla{\rm log}\phi(\mat{x},t) = -\frac{1}{p(\mat{x,t})}\int \nabla G(\mat{x},t;\mat{x}')p_{\rm data}(\mat{x}')d^N\mat{x}'=\mathbb{E}_{ p_t(\mat{x}'|\mat{x})} \left(\frac{\mat{x}-\mat{x}'}{t}\right) \\
        R(\mat{x},t) &= 0
    \end{aligned}
\end{equation}
where $p_t(\mat{x}'|\mat{x})\propto p_{\rm data}(\mat{x}')G(\mat{x},t;\mat{x}')\sim p_{\rm data}(\mat{x}'){\rm exp}(-|\mat{x}-\mat{x}'|^2/(2t))$. Note that $-\mat{v}(\mat{x},t)$ recovers the \textit{score function} in~\citep{Song2019GenerativeMB,song2020score}, \textit{i.e.,} $\nabla_\rvx \log p(\rvx, t)$.

To check that {\bf (C2)} holds, we define $F$ to measure the independence of the final distribution on the initial condition:
\begin{equation}
        F(\mat{x}_1',\mat{x}_2',T)\equiv\int\sqrt{p(\mat{x},T;\mat{x}_1')}\sqrt{p(\mat{x},T;\mat{x}_2')}d^N\mat{x} = {\rm exp}\left(-\frac{|\mat{x}_1-\mat{x}_2|^2}{8T}\right),
\end{equation}
where $F=1$ means independence, and $F=0$ means dependence. We have $\underset{T\to\infty}{{\rm lim}}F(\mat{x}_1',\mat{x}_2',T)\to 1$, implying data-independent priors.

{\bf Example 2: Poisson flow generative models}
We aim to convert the Poisson equation $-(\phi_{tt}+\nabla^2\phi)=p_{\rm data}(\mat{x})\delta(t)$ to a density flow $\frac{\partial p}{\partial t}+\nabla\cdot(p\mat{v})-R=p_{\rm data}(\mat{x})\delta(t)$:
\begin{equation}
    -(\phi_{tt} + \nabla^2\phi) = \frac{(\partial\textcolor{red}{-\phi_t})}{\partial t} + \nabla\cdot\left[\textcolor{red}{-\phi_t}\left(\textcolor{blue}{\frac{\nabla\phi}{\phi_t}}\right)\right] - \textcolor{green}{0}=0\quad \Leftrightarrow\quad  \frac{\partial \textcolor{red}{p}}{\partial t}+\nabla\cdot(\textcolor{red}{p}\textcolor{blue}{\mat{v}})-\textcolor{green}{R}=0
\end{equation}

Comparing the two sides gives:
\begin{equation}\label{eq:poisson-match}
    p = -\phi_t,\quad \mat{v} =\frac{\nabla\phi}{\phi_t},\quad R=0
\end{equation}
Note that we are reinterpreting as time what physicists consider as one of the $N+1$ spatial dimensions. For $N>2$, The solution $\phi$ of the Poisson equation is:
\begin{equation}\label{eq:poisson-phi-solution}
    \phi(\mat{x},t) = \int G(\mat{x},t;\mat{x}')p_{\rm data}(\mat{x}')d^N\mat{x}',\quad G(\mat{x},t;\mat{x}')=\frac{1}{A_N}\frac{1}{(t^2+|\mat{x}-\mat{x}'|^2)^{\frac{N-1}{2}}}
\end{equation}
where $A_N$ is the surface area of a unit $N$-sphere. Combining Eq.~(\ref{eq:poisson-match}) and Eq.~(\ref{eq:poisson-phi-solution}) gives
\begin{equation}
    \begin{aligned}
        p(\mat{x}, t) & = -\phi_t(\mat{x},t) = \int -G_t(\mat{x},t;\mat{x}')p_{\rm data}(\mat{x}')d^N\mat{x}'\\
        &=\frac{N-1}{2A_N}\int \frac{t}{(t^2+|\mat{x}-\mat{x}'|^2)^{\frac{N+1}{2}}}p_{\rm data}(\mat{x}')d^N\mat{x}'  \\
        \mat{v}(\mat{x},t) & = \frac{\nabla\phi(\mat{x},t)}{\phi_t(\mat{x},t)} = \frac{1}{p(\mat{x,t})}\int \nabla G(\mat{x},t;\mat{x}')p_{\rm data}(\mat{x}')d^N\mat{x}'=\mathbb{E}_{ p_t(\mat{x}'|\mat{x})} \left(\frac{\mat{x}-\mat{x}'}{t}\right) \\
        R(\mat{x},t) &= 0
    \end{aligned}
\end{equation}
where $p_t(\mat{x}'|\mat{x})\propto p_{\rm data}(\mat{x}')G(\mat{x},t;\mat{x}')\sim p_{\rm data}(\mat{x}')/(t^2+|\mat{x}-\mat{x}'|^2)^{(N+1)/2}$. Note that $\mat{v}(\mat{x},t)$ recovers the \textit{Poisson fields} in~\citep{xu2022poisson}.

We check that {\bf (C2)} holds. Although $F(\mat{x}_1',\mat{x}_2',t)\equiv\int\sqrt{p(\mat{x},T;\mat{x}_1')}\sqrt{p(\mat{x},T;\mat{x}_2')}d^N\mat{x}$ may not have a closed form for any $t>0$, we notice that for large $T$:
\begin{equation}
    p(\mat{x},T;\mat{x}')\sim \frac{1}{(1+\frac{|\mat{x}-\mat{x}'|^2}{T^2})^{\frac{N+1}{2}}}\approx {\rm exp}(-\frac{(N+1)|\mat{x}-\mat{x}'|^2}{2T^2}),
\end{equation}
so we can show $\underset{T\to\infty}{\rm lim} F(\mat{x}_1',\mat{x}_2',T)\to 1$ similar to the diffusion equation case, implying data-independent priors.

\section{Which physical processes can be converted to GenPhys?}\label{sec:processes}

In this section, we will examine several physical PDEs, rewrite and interpret them as generative models. We will also use the two conditions {\bf (C1)} and {\bf (C2)} to determine whether these PDEs are s-generative. Our main results and illustrations (for a toy two-point distribution) are presented below and  summarized in Table \ref{tab:1} and \ref{tab:2}, with derivation details deferred to Appendix \ref{app:green_function}.

\begin{table}[tbp]
    \centering
    \resizebox{\textwidth}{!}{%
    \begin{tabular}{|c|c|c|c|c|}\hline
     equation  & diffusion equation &  Poisson equation & ideal wave equation & dissipative wave equation   \\\hline
     PDE $\hat{L}\phi=0$  & $\phi_t-\nabla^2 \phi=0$ &  $\phi_{tt}+\nabla^2\phi=0$ & 
     $\phi_{tt}-\nabla^2\phi=0$ & $\phi_{tt}+ 2\epsilon\phi_{t}-\nabla^2\phi = 0$ \\\hline
     rewritten & $\frac{\partial\phi}{\partial t} + \nabla\cdot (\phi(-\nabla{\rm log}\phi))=0$ & $\frac{\partial (-\phi_t)}{\partial t} +  \nabla\cdot((-\phi_t) (\frac{\nabla\phi}{\phi_t}))=0$ & $\frac{\partial (-\phi_t)}{\partial t} + \nabla\cdot((-\phi_t) (-\frac{\nabla\phi}{\phi_t}))=0$ & \scalebox{0.7}{%
    $\frac{\partial(-\phi_t-2\epsilon\phi)}{\partial t} + \nabla\cdot((-\phi_t-2\epsilon\phi) (\frac{\nabla\phi}{\phi_t+2\epsilon\phi}))=0$ }\\\hline
     $p$ & $\phi$ & $-\phi_t$ & $-\phi_t$ &  $-(\phi_t+2\epsilon\phi)$\\\hline
     $\mat{v}$ & $-\nabla{\rm log}\phi$  & $\frac{\nabla\phi}{\phi_t}$  & $-\frac{\nabla\phi}{\phi_t}$  & $\frac{\nabla\phi}{\phi_t+2\epsilon\phi}$ \\\hline
     $R$ & 0 & 0 & 0 & 0 \\\hline
     $G(r,t)$ & $\frac{1}{(4\pi t)^{\frac{N}{2}}}{\rm exp}(-\frac{r^2}{4t})$ & $\frac{1}{(t^2+r^2)^{\frac{N-1}{2}}}$ & $\frac{1}{\sqrt{t^2-r^2}}\Theta(t-r)\ {\rm(2D)}$ & \scalebox{0.7}{$\frac{e^{-\epsilon t}{\rm cosh}(\epsilon\sqrt{t^2-r^2})}{\sqrt{t^2-r^2}}\Theta(t-r)\ {\rm(2D)}$} \\\hline
     $\hat{G}(k,t)$ & ${\rm exp}(-k^2 t)$ & ${\rm exp}(-k t)$ & ${\rm exp}(\pm ikt)$ & \makecell{${\rm exp}(-\epsilon t+i\sqrt{k^2-\epsilon^2}t)\ (k>\epsilon)$ \\ ${\rm exp}(-(\epsilon+\sqrt{k^2-\epsilon^2})t)\ (k\leq \epsilon)$}    \\\hline
     {\bf (C1)} & Yes & Yes & No & Conditionally yes \\\hline
     {\bf (C2)} & Yes & Yes & No & Conditionally yes \\\hline
     \makecell[c]{Illustration \\ $\phi$} & \begin{minipage}{.3\textwidth}
      \includegraphics[width=\linewidth, height=60mm]{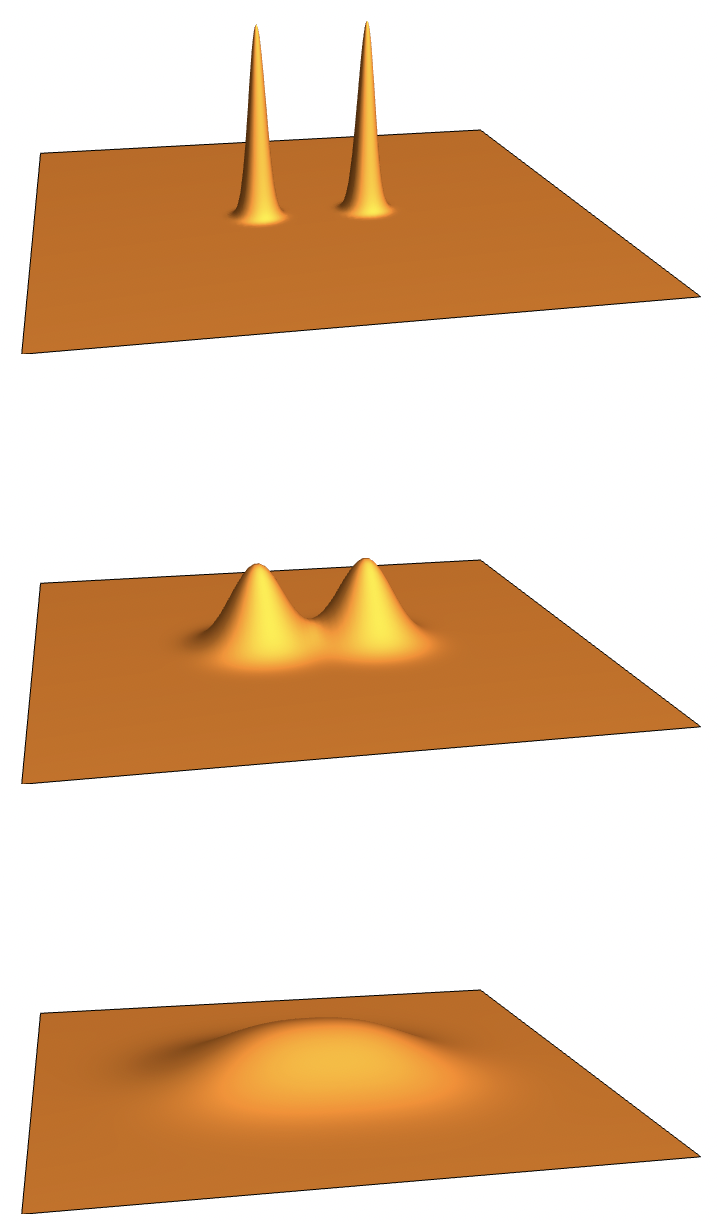}
    \end{minipage}
    & 
    \begin{minipage}{.3\textwidth}
      \includegraphics[width=\linewidth, height=60mm]{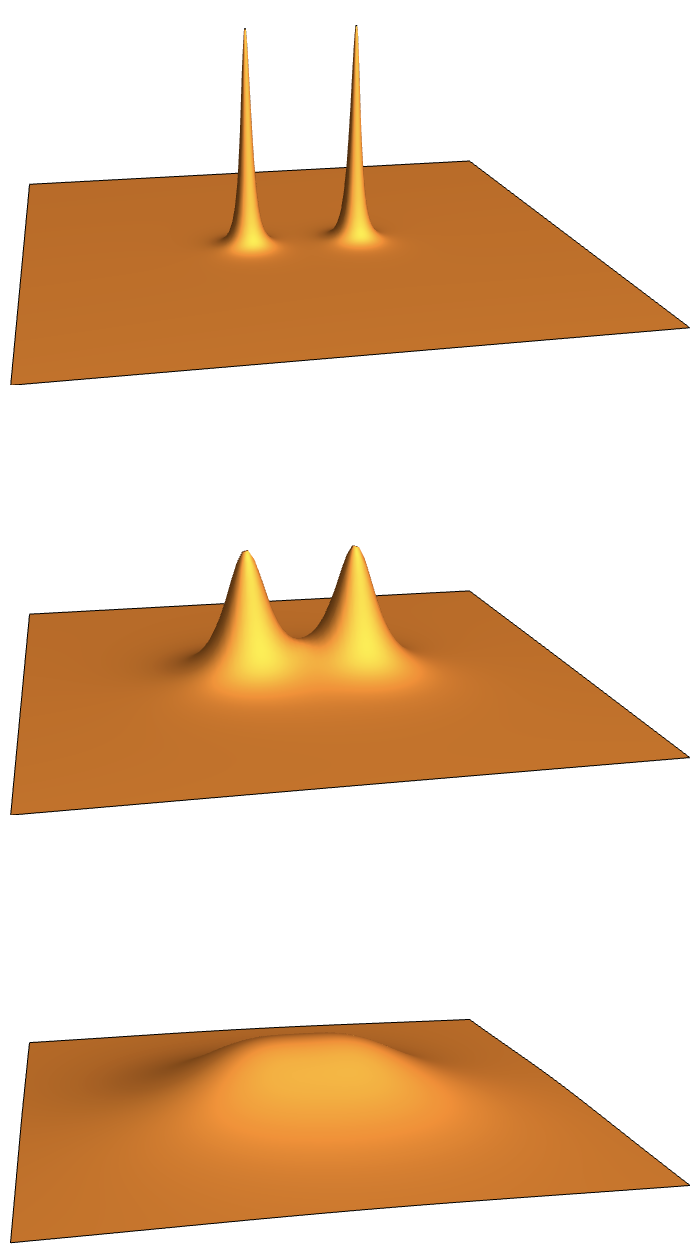}
    \end{minipage}
    &
    \begin{minipage}{.3\textwidth}
      \includegraphics[width=\linewidth, height=60mm]{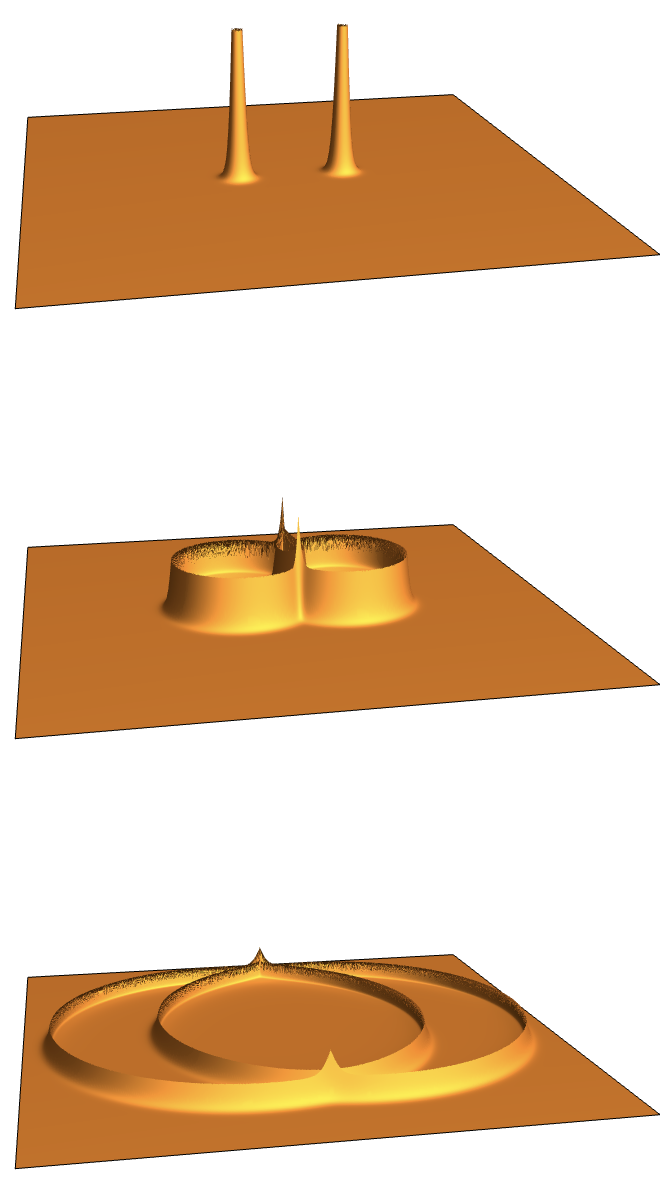}
    \end{minipage}
    & \begin{minipage}{.3\textwidth}
      \includegraphics[width=\linewidth, height=60mm]{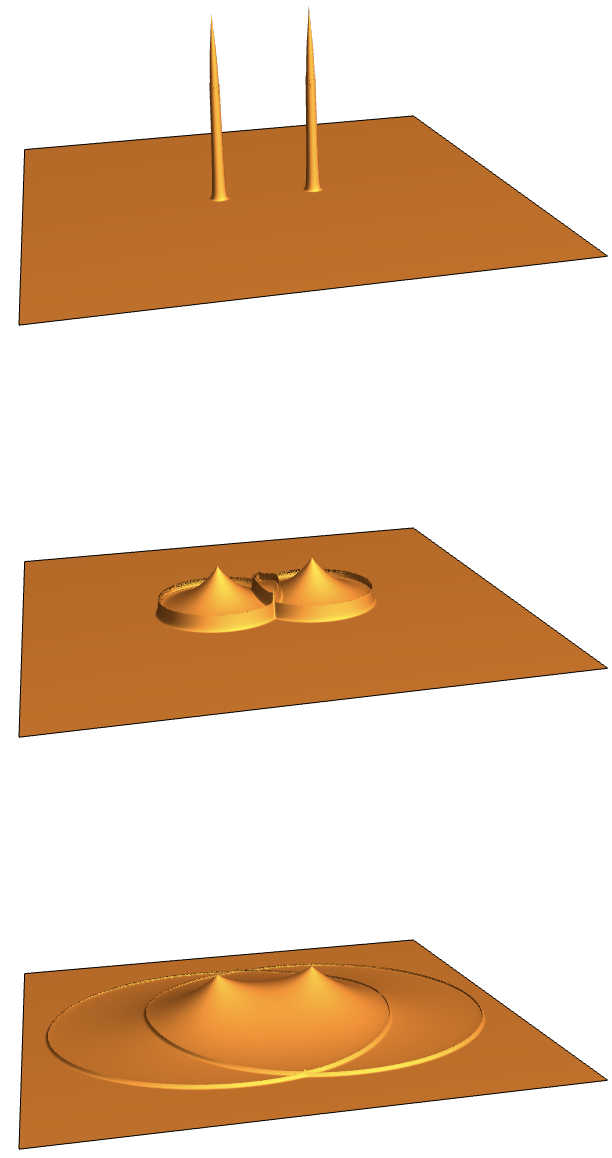}
    \end{minipage}
    \\\hline
     s-generative? & Yes (Diffusion Models) & Yes (Poisson Flow) & No & Conditionally Yes (large $\epsilon$) \\\hline
    \end{tabular}}
    \caption{Connections between physical PDEs and generative models. $r\equiv|\mat{x}-\mat{x}'|$, where $\mat{x}'$ and $\mat{x}$ are the source point and the field point, respectively.}
    \label{tab:1}
\end{table}

{\bf The ideal wave equation (not s-generative)} is  $\phi_{tt}-\nabla^2\phi=0$, describing the propagation of waves of sound, light, etc~\citep{soodak1993wakes}. The wave equation preserves information in the sense that a wave front travels with a constant speed away from the source. Think of a stone dropped into water. Rewritten as $\frac{\partial (-\phi_t)}{\partial t}+\nabla\cdot((-\phi_t)(-\frac{\nabla\phi}{\phi_t}))-0=0$,  we have $p=-\phi_t$, $\mat{v}=-\frac{\nabla\phi}{\phi_t}, R=0$. The Green's function for $N=2$~\footnote{Please refer to Appendix \ref{app:green_function} for general $N$.} is $G(\mat{x},t;\mat{x}')=\Theta(t-r)/\sqrt{t^2-r^2}$ where $\Theta$ is the step function, $r\equiv |\mat{x}-\mat{x}'|$. {\bf (C1)} fails: $\mat{v}$ diverges at the ``wave front" ($r=t$).
{\bf (C2)} fails: the wave front preserves initial information along the way, so the final distribution is dependent on the initial distribution. This is why sound waves are useful for communication. In summary, the ideal wave equation is not s-generative.

{\bf The dissipative wave equation (conditionally s-generative)} is $\phi_{tt}+2\epsilon\phi_t-\nabla^2\phi=0$ where $\epsilon$ is the damping coefficient~\citep{wave_green}. It describes wave propagation with dissipation. Dissipation slows down the wave propagation, leading to behavior ``interpolating" between wave and diffusion:  $\epsilon\to 0$ recovers ideal waves, and $\epsilon\to \infty$ recovers diffusion. Rewritten as $\frac{\partial (-\phi_t-2\epsilon\phi)}{\partial t}+\nabla\cdot((-\phi_t-2\epsilon\phi)(-\frac{\nabla\phi}{\phi_t+2\epsilon\phi}))-0=0$, we have $p=-\phi_t-2\epsilon\phi$, $\mat{v}=-\nabla\phi/(\phi_t+2\epsilon\phi)$, $R=0$. As shown in Table \ref{tab:1} last column, compared to ideal waves, dissipation reduces the magnitude of wave fronts. Behind the wave front, the wave behaves more like diffusion (although with a non-Gaussian kernel). Choosing a large enough $\epsilon$ will make the process quantitatively similar to diffusion, so the two conditions should (approximately) hold. More investigations are needed in the future to determine the exact condition for $\epsilon$, but temporarily we can categorize the dissipative wave equation as conditionally s-generative.

\begin{table}[tbp]
    \centering
    \resizebox{\textwidth}{!}{%
    \begin{tabular}{|c|c|c|c|}\hline
     equation  & Helmholtz equation &  screened Poisson equation (Yukawa) & Schr\"{o}dinger equation  \\\hline
     PDE $\hat{L}\phi=0$  & $\phi_{tt}+\nabla^2\phi+ k_0^2\phi=0$ &  $\phi_{tt}+\nabla^2\phi-m^2\phi=0$ & 
     $i\phi_t+\nabla^2\phi=0$  \\\hline
     Rewritten & $\frac{\partial(-\phi_t)}{\partial t}+\nabla\cdot((-\phi_t)(\frac{\nabla\phi}{\phi_t}))-k_0^2\phi=0$ & $\frac{\partial(-\phi_t)}{\partial t}+\nabla\cdot((-\phi_t)(\frac{\nabla\phi}{\phi_t}))+m^2\phi=0$ & $\frac{\partial|\phi|^2}{\partial t}+\nabla\cdot(|\phi|^2(2{\rm Im}\nabla {\rm log}\phi))=0$  \\\hline
     $p$ & $-\phi_t$ & $-\phi_t$ & $|\phi|^2$  \\\hline
     $\mat{v}$ & $\frac{\nabla\phi}{\phi_t}$ & $\frac{\nabla\phi}{\phi_t}$ & $2{\rm Im}\nabla{\rm log}\phi$  \\\hline
     $R$ & $k_0^2\phi$ & $-m^2\phi$ & 0  \\\hline
     $G(r,t)$ & $(\frac{k_0}{\sqrt{t^2+r^2}})^{\frac{N-1}{2}}H^{(1)}_{\frac{N-1}{2}}(k_0\sqrt{t^2+r^2})$ & $(\frac{m}{\sqrt{t^2+r^2}})^{\frac{N-1}{2}}K_{\frac{N-1}{2}}(m\sqrt{t^2+r^2})$ & $\frac{1}{(4\pi it)^{\frac{N}{2}}}{\rm exp}(\frac{ir^2}{4t})$ \\\hline
     $\hat{G}(k,t)$ & \makecell{${\rm exp}(-i\sqrt{k_0^2-k^2}t)\ (k\leq k_0)$ \\ ${\rm exp}(-\sqrt{k^2-k_0^2}t)\ (k>k_0)$} & ${\rm exp}(-\sqrt{k^2+m^2}t)$ & ${\rm exp}(ik^2t)$ \\\hline
     {\bf (C1)} & Conditional yes & Yes & No \\\hline
     {\bf (C2)} & Conditional Yes & Yes & No \\\hline
     \makecell[c]{Illustration \\ $\phi$ or ${\rm Re}\phi$ } & \begin{minipage}{.3\textwidth}
      \includegraphics[width=\linewidth, height=60mm]{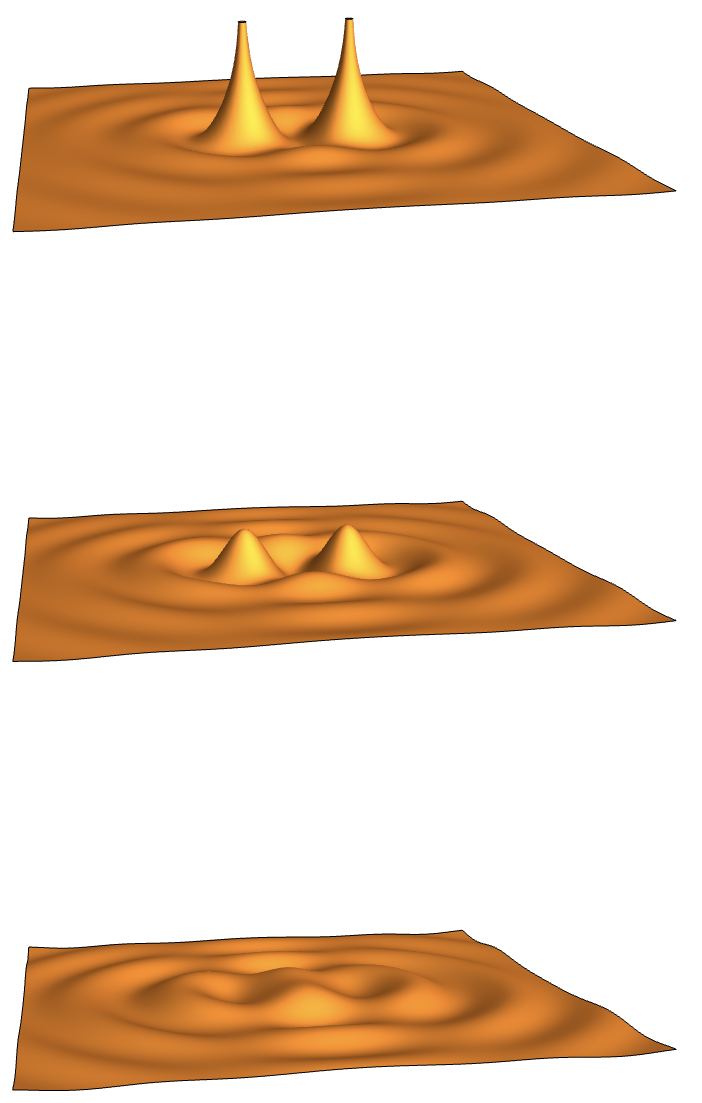}
    \end{minipage}
    & 
    \begin{minipage}{.3\textwidth}
      \includegraphics[width=\linewidth, height=60mm]{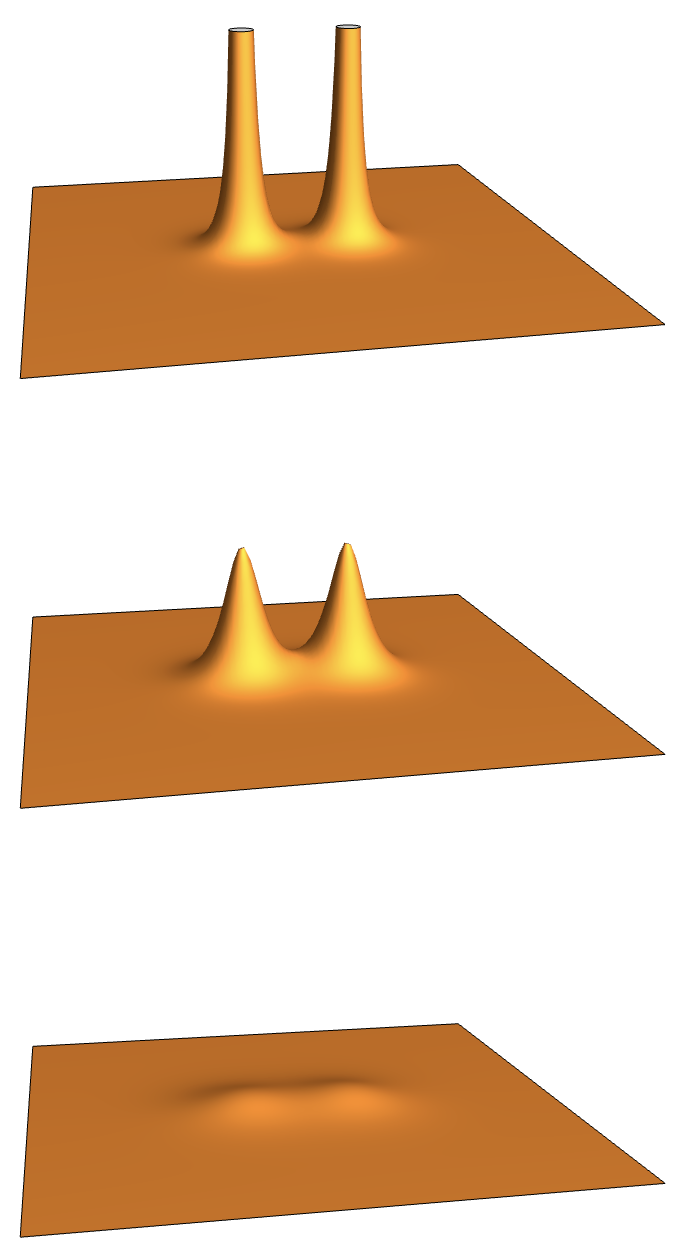}
    \end{minipage}
    &
    \begin{minipage}{.3\textwidth}
      \includegraphics[width=\linewidth, height=60mm]{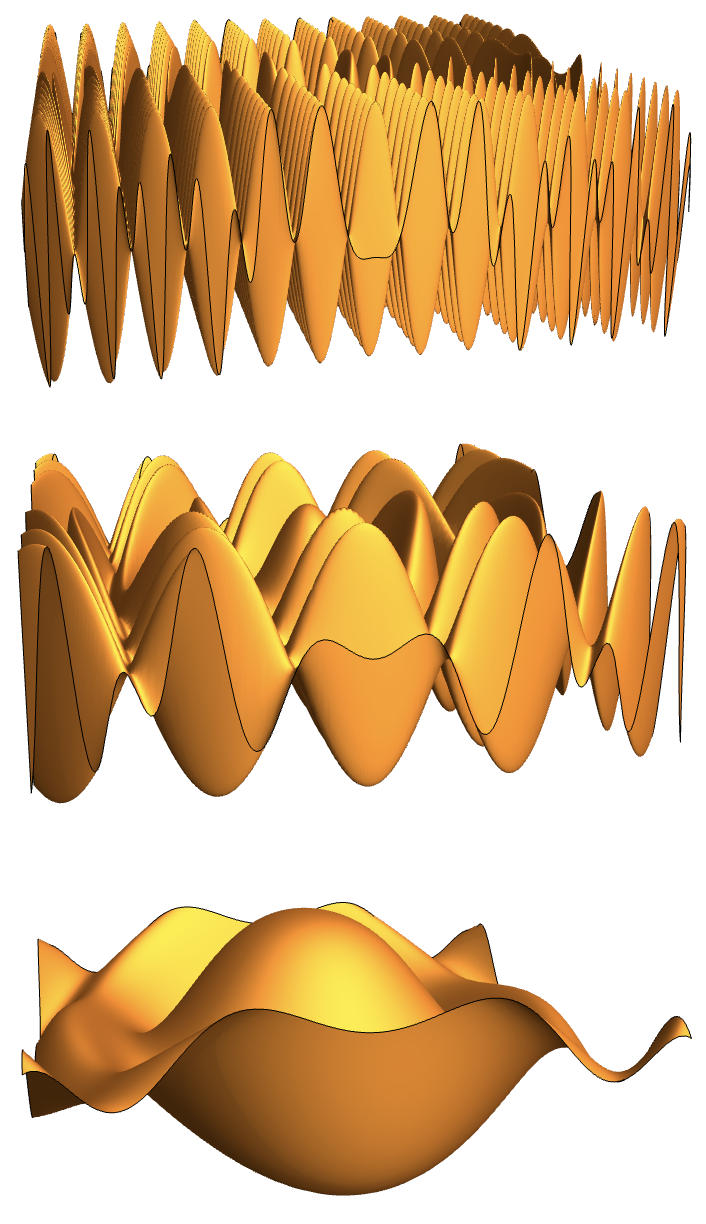}
    \end{minipage}
    \\\hline
    s-generative? & Conditionally yes (small $k$) & Yes & No \\\hline
    \end{tabular}}
    \caption{ (Continuing Table \ref{tab:1}) Connections between physical PDEs and generative models. $r\equiv|\mat{x}-\mat{x}'|$, where $\mat{x}'$ and $\mat{x}$ are the source point and the field point, respectively.}
    \label{tab:2}
\end{table}

{\bf The Helmholtz equation (conditionally s-generative)} $(\nabla_{\tilde{\mat{x}}}^2+k_0^2)\phi=0$ is the single-frequency wave equation, which explains the wave-like behavior (shown in Table \ref{tab:2}) in its Green's function. Think of water ripples driven by a periodic source (instead of a one-shot perturbation, e.g., a stone dropped into water). Similar to the Poisson equation, we set $\tilde{\mat{x}}=[t,\mat{x}]$. Rewritten as $\frac{\partial (-\phi_t)}{\partial t}+\nabla\cdot \left[(-\phi_t)(\frac{\nabla\phi}{\phi_t})\right]-k_0^2\phi=0$, we can match $p=-\phi_t$, $\mat{v}=\nabla\phi/\phi_t$ and $R=k_0^2\phi$. {\bf (C1)} conditionally holds. $p=-\phi_t$ is in general not a proper density distribution because it can be negative. That said, $p$ remains positive for the local region $r\equiv|\mat{x}-\mat{x}'|\ll\frac{2\pi}{k_0}$, so a small enough $k_0$ can make {\bf (C1)} hold. In fact, $k_0\to 0$ recovers the Poisson equation, so a small enough $k_0$ should also (at least approximately) make {\bf (C1)} \& {\bf (C2)} hold. More investigations are needed in the future to determine the exact condition for $k_0$, but temporarily we categorize the Helmholtz equation as conditionally s-generative.

{\bf The screened Poisson equation, a.k.a. Yukawa (s-generative)} $(\nabla_{\tilde{\mat{x}}}^2-m^2)\phi=0$, where $m$ expresses the ``screening". Compared to the Poisson potential (the solution of the Poisson equation), the screened Poisson potential is more short-ranged with a length scale $m^{-1}$, due to screening effects. When $N=3$, $\phi$ is the well-known Yukawa potential~\citep{yukawa1937interaction}. Note that $m=0$ recovers the Poisson equation. As for the Poisson equation, we set $\tilde{\mat{x}}=[t,\mat{x}]$. Rewritten as $\frac{\partial -\phi_t}{\partial t}+\nabla\cdot\left[(-\phi_t)(\frac{\nabla\phi}{\phi_t})\right]+m^2\phi=0$, we can match  $p=-\phi_t$, $\mat{v}=\nabla\phi/\phi_t$, $R=m^2\phi$. The Green's function of the screened Poisson equation is $G(\mat{x},t;\mat{x}')=(\frac{m}{\sqrt{t^2+r^2}})^{\frac{N-1}{2}}K_{\frac{N-1}{2}}(m\sqrt{t^2+r^2})$, where $K$ is the modified Bessel function of the second kind. The solution is qualitatively similar to that of the Poisson equation, but decreases faster with distance. We can check that {\bf (C1)} \& {\bf (C2)} hold.

{\bf The Schr\"odinger equation (not s-generative)} is $i\phi_t = -\nabla^2 \phi + V(\mat{x}) \phi$, where $\phi$ is the (complex-valued) wave function. It describes the evolution of a quantum particle~\citep{griffiths2018introduction}. The Schr\"{o}dinger equation describes the wave nature of the particles, based on the idea of wave-particle duality. For $V(\mat{x})=0$, the free particle PDE implies $\frac{\partial |\phi|^2}{\partial t}+\nabla\cdot\left[|\phi|^2(2{\rm Im}\nabla{\rm log}\phi)\right]=0$ (see Appendix \ref{app:green_function} for details), so $p=|\phi|^2$, $\mat{v}=2\ {\rm Im}\nabla{\rm log}\phi$ and $R=0$. The Green's function $G(\mat{x},t;\mat{x}')=\frac{1}{(4it)^{N/2}}{\rm exp}(-\frac{|\mat{x}-\mat{x}'|^2}{4it})$.  {\bf (C1)} fails. Although $p\equiv |\phi|^2$ is a probability distribution, it may have zeros due to interference, causing divergence of $\mat{v}$. {\bf (C2)} fails. $p(\mat{x},T)$ oscillates restlessly even for large $T$ and depends on the initial distribution. In summary, the Schr\"odinger equation is not s-generative. However, to have the finial distribution independent of the initial distribution, it is possible to consider the subsystem quantum dynamics under the Schr\"odinger evolution such that it thermalizes, or reaches a steady state in the open quantum system formulation.

{\bf Remark: Interpolating between GenPhys gives a spectrum of GenPhys} One can get a spectrum of GenPhys by interpolating between two GenPhys. For example, to interpolate between DM and PFGM, one can study the PDE $a\phi_{tt} - b\phi_t + \nabla^2\phi=0\ (a\geq0,b\geq0)$. Note that $(a,b)=(1,0)$ and $(a,b)=(0,1)$ recover PFGM and DM, respectively. The PDE can be rewritten as $\frac{\partial (-a\phi_t+b\phi)}{\partial t}=-\nabla\cdot((-a\phi_t+b\phi)(\frac{\nabla\phi}{a\phi_t-b\phi}))$, so $p\equiv -a\phi_t+b\phi$ and $\mat{v}\equiv\frac{\nabla\phi}{a\phi_t-b\phi}$. We can check that both {\bf (C1)} and {\bf (C2)} hold, so the intepolated GenPhys is also valid. However, the Green's function $\phi$ may not admit a closed from for general $a, b$ (see Appendix \ref{app:interpolating_DM_PFGM}).

\section{Dispersion relation as a criterion}\label{sec:dispersive-relation}

Checking {\bf (C2)} is  case-dependent and usually math heavy, as we demonstrate in Appendix \ref{app:green_function}. In this section, we show that {\bf (C2)} can boil down to a restriction on the dispersion relation of PDEs, so  the dispersion relation is a convenient and principled tool to validate and even construct GenPhys.

{\bf The dispersion relation} relates the wavenumber $k$ of a wave to its frequency $\omega$. All linear PDEs $\hat{L}\phi(\mat{x},t)=0$ have wave solutions: $\phi(\mat{x},t)\propto{\rm exp}(-i\omega t){\rm exp}(i\mat{k}\cdot\mat{x})$. Substituting the wave ansatz into the PDE gives the dispersion relation $\omega(\mat{k})$. For example, the diffusion equation $\phi_t - \nabla^2\phi=0$ has $\omega=-ik^2 (k\equiv|\mat{k}|)$, where $w$ is purely imaginary, so the time factor ${\rm exp}(-i\omega t)\sim {\rm exp}(-k^2t)$ decays in time. In contrast, the wave equation $\phi_{tt}-\nabla^2\phi=0$ has $\omega(k)=\pm k$, where $\omega$ is real, so the time factor ${\rm exp}(\pm ikt)$ oscillates in time without any decay. 
\begin{table}[htbp]
    \centering
    \caption{Dispersion relation $\omega(k)$ of physical PDEs.}\label{tab:dispersion}
    \begin{tabular}{cccc}
    \toprule
     Physics  &  PDE & Dispersion Relation & s-generative? \\
     \midrule
     Diffusion    &  $\phi_t-\nabla^2\phi=0$  & $\omega=-ik^2$ &  Yes \\
     Poisson & $\phi_{tt}+\nabla^2\phi=0$ & $\omega=\pm ik$ & Yes \\
     Ideal Wave & $\phi_{tt}-\nabla^2\phi=0$ & $\omega=\pm k$ & No \\
     Dissipative wave & $\phi_{tt}+2\epsilon\phi_t-\nabla^2\phi=0$ & $\omega=\begin{cases}
    i(-\epsilon\pm\sqrt{\epsilon^2-k^2})&k\leq\epsilon \\
    i\epsilon\pm\sqrt{k^2-\epsilon^2}&k>\epsilon
\end{cases}$ & \makecell{Conditionally \\ Yes (large $\epsilon$)} \\
    Helmholtz & $\phi_{tt}+\nabla^2\phi+k_0^2\phi=0$ & $\omega=\begin{cases}
    \pm\sqrt{k_0^2-k^2}&k\leq k_0 \\
    \pm i\sqrt{k^2-k_0^2}&k>k_0
\end{cases}$ & \makecell{Conditionally \\ Yes (small $k_0^2$)} \\
    Screened Poisson & $\phi_{tt}+\nabla^2\phi-m^2\phi=0$ & $\omega=\pm i\sqrt{k^2+m^2}$ & Yes \\
    Schr\"{o}dinger & $i\phi_t+\nabla^2\phi=0$ & $\omega=k^2$ & No \\
    \bottomrule
    \end{tabular}
\end{table}

{\bf A PDE is smooth = A dispersion criterion}
In Table \ref{tab:dispersion}, we list the dispersion relations $\omega(k)$ of several PDEs discussed above. The last column lists whether the PDE is s-generative. There is a perfect correlation between the dispersion relation and being s-generative: (1) s-generative equations have pure imaginary $\omega$ for all $k$, (2) conditionally s-generative equations have pure imaginary $\omega$ for some $k$ range, and (3) not s-generative equations have $k$ range with pure imaginary $\omega$. We can prove (see Appendix \ref{app:C2-dispersion}) that {\bf (C2)} is in fact equivalent to the following condition:
\begin{equation}\label{eq:C2-dispersion}
\boxed{
    {\rm Im}\ \omega(k) < {\rm Im}\ \omega(0),\quad {\rm for\ all\ } k>0
}
\end{equation}
Note that although dispersion relations may have multiple branches, we only require one branch to satisfy Eq.~(\ref{eq:C2-dispersion}). For example, dispersion relations of the Poisson equation have two branches $\omega=\pm ik$, one of which ($\omega=-ik$) satisfies Eq.~(\ref{eq:C2-dispersion}). Intuitively, {\bf (C2)} requires that details of the initial distribution smooth out as time elapses, which means that high-frequency modes should decay faster than low-frequency ones. Eq.~(\ref{eq:C2-dispersion}) states that non-zero frequency modes should decay faster than the zero frequency mode. If we define $\hat{p}(\mat{k})$ as the spatial Fourier transform of $p(\mat{x})$, this condition simply means that spatial oscillations with $k\equiv|\mat{k}|>0$ decay faster than the total probability/mass $\hat{p}(\mat{k}=0, t)=\int p(\mat{x}, t)d^N\mat{x}$.

{\bf Constructing generative models via dispersion relations}  We can now construct GenPhys simply by constructing PDEs whose dispersion relations satisfy the restriction Eq.~(\ref{eq:C2-dispersion}). For example, four PDEs listed in Table \ref{tab:new-gm} satisfy the dispersion relation restriction, so the corresponding GenPhys are worth further investigating in the future.

\begin{table}[htbp]
    \centering
    \caption{The dispersion relation suggests new GenPhys}
    \begin{tabular}{cccc}
    \toprule
     Physics  &  PDE & Dispersion Relation \\
     \midrule
    Mixed diffusion Poisson& $a\phi_{tt}-b\phi_t+\nabla^2\phi=0\ (a>0,b>0)$ & $\omega = \frac{i}{2a}(b\pm\sqrt{b^2+4ak^2})$ \\
    Fractional diffusion & $\phi_t+(-\Delta)^\beta\phi=0\ (\beta>0)$ & $\omega=-ik^{2\beta}$ \\
    Third-order ``diffusion" & $\phi_{ttt}-\Delta u=0$ & $\omega=(-i,e^{i\frac{\pi}{6}},e^{i\frac{5\pi}{6}})k^{\frac{2}{3}}$ \\
    Elasticity (Biharmonic) & $\phi_t+\nabla^2\nabla^2 \phi=0$ & $\omega=-ik^4$ \\
    \bottomrule
    \end{tabular}
    \label{tab:new-gm}
\end{table}

\def\onedot{$\mathsurround0pt\ldotp$}
\def\ie{\emph{i.e}\onedot, }
\def\rvx{{\mathbf{x}}}
\section{Related work}\label{sec:related_works}

A line of prominent physics-inspired generative models in the machine learning field traces back to the energy-based models~(EBM)~\citep{LeCun2006ATO}. EBM casts the generative process as finding low energy states through the simulation of Langevin dynamics~\citep{Parisi1981CorrelationFA}. To bypass the costly MCMC sampling during the EBM training, score-matching~\citep{Hyvrinen2007SomeEO,Vincent2011ACB} estimates the gradient of the energy function (score) rather than modeling the energy directly. To address the instability of score-matching on low-dimensional data manifolds, \cite{Song2019GenerativeMB} combined annealed Langevin dynamics with the scores of perturbed distributions with different levels of Gaussian noise. Later, \cite{song2020score} showed that both the annealed Langevin dynamics score-matching and diffusion models with discrete Markov chains~\citep{sohl2015deep, ho2020denoising} can be integrated into a continuous Stochastic Differential Equation~(SDE) framework, where the generative process is equivalent to reversing a fixed forward diffusion process. This framework has been applied to various tasks, including text-to-image generation~\citep{Rombach2021HighResolutionIS, Saharia2022PhotorealisticTD}, 3D generation~\citep{Poole2022DreamFusionTU, Zeng2022LIONLP} and molecule generation~\citep{Hoogeboom2022EquivariantDF}.

The more recent Poisson Flow Generation Models~(PFGM)~\citep{xu2022poisson, Xu2023PFGMUT} arises from electrostatics and rivals diffusion models in image generation. PFGM regards the data as charges and performs generative modeling by evolving the samples along electric field lines in an augmented space. This approach has been shown to exhibit better sample quality and greater robustness than diffusion models. Another physics-inspired work~\citep{Rissanen2022GenerativeMW} generates samples by iteratively inverting the heat equation over the 2D plane of the image.

\section{Conclusions and discussions}\label{sec:conclusion}

We present a framework to convert any s-generative partial differential equations to generative models. We focus on equations with physical meaning, calling the resulting generative models {\bf Gen}erative Models from {\bf Phys}ical Processes ({\bf GenPhys}). Besides the two existing GenPhys, diffusion models and Poisson flow generative models, our framework allows automatic generation of more GenPhys. The goal of this paper is to point out the existence of more GenPhys; the next step will be analyzing and testing their behavior in practice. The million dollar question is: can those new GenPhys beat the existing ones in terms of theoretical guarantee and practical performance?

Our framework identifies a way to construct generative models from physical processes, but there are still many great opportunities not covered by this framework. For example: (1) We only leveraged smoothing PDEs, but cannot take full advantage of non-smoothing PDEs. There exists non s-generative model which can also provide useful generative modeling, such as the case in quantum machine learning with dynamics based on the Schr\"odinger equation  and quantum circuits. (2) We only discussed linear PDEs, but nonlinear PDEs may also inspire generative models, e.g., the Navier-Stokes equation, the reaction diffusion equation and the Bose-Einstein condensation etc.  It would be particularly interesting to see if these nonlinear physical phenomena can inspire generative models, since they are prevalent in our physical world. (3) We only discussed PDEs with translational or rotational symmetry with analytical accessibility of the Green's function, but generic PDEs without such symmetries may offer additional power and flexibility. For example, the imaginary time evolution of a Schr\"odinger equation with position-dependent potential gives rise to birth/death process in the density flow. (4) We only considered time-independent PDEs, but time-dependent PDEs offer a more general framework. This opens up connections to a broader class of dynamical processes in nature, such as non-equilibrium dynamics, quench experiments and annealing. For example, it is also known that adiabatic quantum dynamics can achieve universal quantum computation~\citep{aharonov2008adiabatic} which can be leveraged for generic probability modeling.

\section*{Acknowledgement} 
We would like to thank Zhuo Chen, Hongye Hu and Catherine Liang for helpful discussions. YX and TJ acknowledge support from MIT-DSTA Singapore collaboration, from NSF Expeditions grant (award
1918839) “Understanding the World Through Code”, and
from MIT-IBM Grand Challenge project. ZL and MT are supported by the Foundational Questions
Institute and the Rothberg Family Fund for Cognitive Science. DL is supported by the U.S. Department of Energy, Office of Science, National Quantum Information Science Research Centers, Co-design Center for Quantum Advantage (C2QA) under contract number DE-SC0012704.
ZL, MT and DL are supported by IAIFI through NSF grant PHY-2019786. 

\clearpage
\bibliography{iclr2023_workshop}
\bibliographystyle{iclr2023_workshop}
\clearpage
{\huge Appendix}
\appendix
\section{Green's function review}\label{app:green_function}

\subsection{Math basics}

{\bf Fourier transformation} The Green's functions of many PDEs can be found with Fourier transforms. For a scalar function $\phi(\mat{x},t),\ \mat{x}\in\mathbb{R}^N$, we define the Fourier transformation
\begin{equation}\label{eq:fourier}
    \tilde{\phi}(\mat{k}, t) = \mathcal{F}[\phi] \equiv \int \phi(\mat{x},t)e^{-i\mat{k}\cdot\mat{x}} d^N\mat{x},
\end{equation}
and the inverse Fourier transformation 
\begin{equation}\label{eq:inverse-fourier}
    \phi(\mat{x}, t) = \mathcal{F}^{-1}[\tilde{\phi}] \equiv  \frac{1}{(2\pi)^{N}}\int \tilde{\phi}(\mat{k},t)e^{i\mat{k}\cdot\mat{x}} d^N\mat{k}.
\end{equation}

One nice property of Fourier transformation is that derivatives in $\mat{x}$ space becomes multiplication in $\mat{k}$ space, i.e.,
\begin{equation}
    \mathcal{F}[\nabla\phi] = -i\mat{k}\tilde{\phi},\  \mathcal{F}[\nabla^2\phi] = -|\mat{k}|^2\tilde{\phi}
\end{equation}
So solving the PDE in the $\mat{k}$ space can be much simpler than in the $\mat{x}$ space. That said, transforming $\tilde{\phi}(\mat{k},t)$ back to $\phi(\mat{x},t)$ via inverse Fourier transformation may be hard.

{\bf Volume and Surface in high dimensions}
The $n$-dimensional unit ball $\mathcal{B}_N = \{\mat{x}:\sum_{i=1}^N x_i^2\leq 1\}$ has volume  $V_N=\frac{\pi^{\frac{n}{2}}}{\Gamma(\frac{n+1}{2})}$ and surface area $A_{N-1}=nV_n=\frac{2\pi^{\frac{n}{2}}}{\Gamma(\frac{n}{2})}$. The $N$-dimensional volume element $d^N\mat{x}$ expressed in spherical coordinates is
\begin{equation}
    d^N\mat{x} = r^{N-1}dr\prod_{i=1}^{N-2}\left[({\rm sin}\theta_i)^{N-1-i}\ d\theta_i\right]d\varphi
\end{equation}
where $r\equiv|\mat{x}|$, $\theta_i\in[0,\pi]$, for $i=1,\cdots,N-2$, $\varphi\in[0,2\pi)$. Usually one is interested in calculating the integral $\int f(\mat{x}) d^N\mat{x}$, so if $f(\mat{x})$ has axial symmetries except for $\theta_1$, i.e., $f(\mat{x})$ is independent of $\theta_i (i\geq 2)$ and $\phi$, the $\mathbb{R}^N$ space can be sliced into ``rings" based on $\theta_1$ and $r$, and the ring  at $(\theta_1\to \theta_1+d\theta_1, r\to r+dr)$ has the volume
\begin{equation}\label{eq:ring_volume}
    d^N\mat{x} = 2\pi A_{N-2}(r{\rm sin\theta_1})^{N-2}rd\theta_1dr.
\end{equation}

\subsection{Diffusion equation}

The Green's function $G(\mat{x},t)$ of the diffusion equation for $(\mat{x}\in\mathbb{R}^N)$ satisfies:
\begin{equation}
    G_t - \nabla^2G = \delta(\mat{x})\delta(t),
\end{equation}
where for simplicity, we have assumed the source point $\mat{x}'=\mat{0}$. Using Fourier transformation Eq.~(\ref{eq:fourier}), the transformed PDE becomes:
\begin{equation}
    \tilde{G}_t + |\mat{k}|^2 \tilde{G} = \delta(t),
\end{equation}
which is equivalent to
\begin{equation}
    \tilde{G}_t + |\mat{k}|^2 \tilde{G} = 0\ (t>0),\ \tilde{G}(\mat{k},0) = 1.
\end{equation}
This initial value problem has the solution
\begin{equation}
    \tilde{G}(\mat{k},t) = {\rm exp}(-|\mat{k}|^2 t)
\end{equation}
which is a Gaussian distribution in $\mat{k}$ space. Now transforming back to $\mat{x}$ space:

\begin{equation}
\begin{aligned}
    G(\mat{x},t) &= \mathcal{F}^{-1}[\tilde{G}] \\ 
    &=\frac{1}{(2\pi)^N}\int_{\mat{k}} {\rm exp}(-|\mat{k}|^2 t){\rm exp}(i\mat{k}\cdot\mat{x})d^N\mat{k} \\
    &=\frac{1}{(2\pi)^N}\int_{\mat{k}} {\rm exp}(-k^2 t){\rm exp}(ikx{\rm cos}\theta)A_{N-2}(k{\rm sin}\theta)^{N-2}kdkd\theta\quad (k\equiv|\mat{k}|, x\equiv|\mat{x}|)\quad {\rm (invoke\  Eq.~\ref{eq:ring_volume})} \\
    &=\frac{A_{N-2}}{(2\pi)^N}\int_{k=0}^\infty dk k^{N-1}{\rm exp}(-k^2 t)\int_{\theta=0}^\pi {\rm exp}(ikx{\rm cos}\theta)({\rm sin}\theta)^{N-2}d\theta,
\end{aligned}
\end{equation}
and the integral of $\theta$ is:
\begin{equation}
    \int_{\theta=0}^\pi {\rm exp}(ikx{\rm cos}\theta)({\rm sin}\theta)^{N-2}d\theta=\sqrt{\pi}\Gamma(\frac{N-1}{2}){}_0\tilde{\Gamma}_1(\frac{N}{2}, -\frac{(kx)^2}{4}),
\end{equation}
where ${}_0\tilde{\Gamma}_1(b,z)$ is a regularized hypergeometric function. So
\begin{equation}
    \begin{aligned}
        G(\mat{x},t)&=\frac{A_{N-2}\sqrt{\pi}\Gamma(\frac{N-1}{2})}{(2\pi)^N}(\int_{k=0}^\infty k^{N-1}{\rm exp}(-k^2 t) {}_0\tilde{\Gamma}_1(\frac{N}{2}, -\frac{(kx)^2}{4})) \\
        & = \frac{A_{N-2}\sqrt{\pi}\Gamma(\frac{N-1}{2})}{(2\pi)^N} (\frac{1}{2}{\rm exp}(-\frac{x^2}{4t})t^{\frac{N}{2}}) \\
        & = \frac{1}{(4\pi t)^{\frac{N}{2}}}{\rm exp}(-\frac{r^2}{4t}) \\
        & = \frac{1}{(4\pi t)^{\frac{N}{2}}}{\rm exp}(-\frac{|\mat{x}|^2}{4t})
    \end{aligned}
\end{equation}
which is a Gaussian distribution with zero mean and variance $2t$. In fact, there is a much simpler way to compute the inverse Fourier transform, by noticing that the components of $k$ are separable:

\begin{equation}
\begin{aligned}
    G(\mat{x},t) &= \mathcal{F}^{-1}[\tilde{G}] \\ 
    &=\frac{1}{(2\pi)^N}\int_{\mat{k}} {\rm exp}(-|\mat{k}|^2 t){\rm exp}(i\mat{k}\cdot\mat{x})d^N\mat{k} \\
    &=\prod_{i=1}^N(\frac{1}{2\pi}\int_{-\infty}^\infty {\rm exp}(-k_i^2t){\rm exp}(ik_ix_i)dk_i)  \\
    &=\prod_{i=1}^N(\frac{{\rm exp}(-\frac{x_i^2}{4t})}{2\pi}\int_{-\infty}^\infty {\rm exp}(-t(k_i-\frac{ix_i}{2t})^2)dk_i)  \\
    & =\prod_{i=1}^N(\frac{{\rm exp}(-\frac{x_i^2}{4t})}{2\pi}\sqrt{\frac{\pi}{t}})  \\
    &=\frac{1}{(4\pi t)^{\frac{N}{2}}} {\rm exp}(-\frac{|\mat{x}|^2}{4t})
\end{aligned}
\end{equation}

However, the more difficult derivation is more general, and is useful when we attempt to interpolate between DMs and PFGMs (see Appendix \ref{app:interpolating_DM_PFGM}). For general $\mat{x}'$, we thus have
\begin{equation}
    G(\mat{x},t;\mat{x}')=\frac{1}{(4\pi t)^{\frac{N}{2}}}{\rm exp}(-\frac{|\mat{x}-\mat{x}'|^2}{4t})
\end{equation}

For general data distribution $p_{\rm data}(\mat{x})$, we have $\phi(\mat{x},t)=\int p_{\rm data}(\mat{x}')G(\mat{x},t;\mat{x}')d^N\mat{x}'$. We now check the three conditions:

{\bf (C1)} holds. $p(\mat{x},t;\mat{x}')=\phi(\mat{x},t;\mat{x}')\geq 0$.

{\bf (C2)} holds. We have
\begin{equation}
        F(\mat{x}_1',\mat{x}_2',T)\equiv\int\sqrt{p(\mat{x},T;\mat{x}_1')}\sqrt{p(\mat{x},T;\mat{x}_2')}d^N\mat{x} = {\rm exp}(-\frac{|\mat{x}_1-\mat{x}_2|^2}{8T}),
\end{equation}
so $\underset{T\to\infty}{{\rm lim}}F(\mat{x}_1',\mat{x}_2',T)\to 1$, implying data-independent priors.

\subsection{Poisson equation}

The Green's function $\phi(\mat{x})$ of the Poisson equation 
for $(\mat{x}\in\mathbb{R}^N)$ satisfies:
\begin{equation}
    \nabla^2G = \delta(\mat{x})
\end{equation}
where for simplicity we have assumed the source point $\mat{x}'=\mat{0}$. Note that (so far) $G(\mat{x})$ does not depend on time $t$, since Poisson equation physically describes steady states (which are time independent). To bake in the notion of time $t$,  PFGM~\citep{xu2022poisson} augments $\mat{x}\in\mathbb{R}$ to be $ \tilde{\mat{x}}\equiv (\mat{x}, t)\in\mathbb{R}^{N+1}$~\footnote{The original PFGM~\citep{xu2022poisson} uses $z$ instead of $t$.}, where $\mat{x}$ is the data point, and $t$ is the augmented dimension. By splitting $\mat{x}$ and $t$ explicitly, the Poisson equation becomes:
\begin{equation}
    G_{tt} + \nabla^2G = 0,\quad G(\mat{x},0)=\delta(\mat{x}).
\end{equation}
Using Fourier transformation Eq.~(\ref{eq:fourier}) , the transformed PDE becomes:
\begin{equation}
     \tilde{G}_{tt} - |\mat{k}|^2 \tilde{G}=0,\quad \tilde{G}_t(\mat{k},0) = 1,
\end{equation}
whose general solution is $\tilde{G}=\frac{1}{|\mat{k}|}(A{\rm exp}(-|\mat{k}|t)+B{\rm exp}(|\mat{k}|t))$ with $-A+B=1$. Considering the free boundary condition, i.e., as $t\to\infty, \tilde{G}\to 0$, we have $A=-1, B=0$. So $\tilde{G}=-\frac{1}{|\mat{k}|}{\rm exp}(-|\mat{k}|t)$. Now transforming back to $\mat{x}$ space:

\begin{equation}
\begin{aligned}
    G(\mat{x},t) &= \mathcal{F}^{-1}[\tilde{G}] \\ 
    &=\frac{1}{(2\pi)^N}\int_{\mat{k}} \frac{1}{|\mat{k}|}{\rm exp}(-|\mat{k}| t){\rm exp}(i\mat{k}\cdot\mat{x})d^N\mat{k} \\
    &=\frac{1}{(2\pi)^N}\int_{\mat{k}} \frac{1}{k}{\rm exp}(-k t){\rm exp}(ikx{\rm cos}\theta)A_{N-2}(k{\rm sin}\theta)^{N-2}kdkd\theta\quad (k\equiv|\mat{k}|, x\equiv|\mat{x}|)\quad {\rm (invoke\  Eq.~\ref{eq:ring_volume})} \\
    &=\frac{A_{N-2}}{(2\pi)^N}\int_{k=0}^\infty dk k^{N-2}{\rm exp}(-k t)\int_{\theta=0}^\pi {\rm exp}(ikx{\rm cos}\theta)({\rm sin}\theta)^{N-2}d\theta, \\
    &=\frac{A_{N-2}\sqrt{\pi}\Gamma(\frac{N-1}{2})}{(2\pi)^N}(\int_{k=0}^\infty k^{N-2}{\rm exp}(-k t) {}_0\tilde{\Gamma}_1(\frac{N}{2}, -\frac{(kx)^2}{4})) \\
    & = \frac{2}{(4\pi)^{\frac{N}{2}}} (\frac{2^{N-2}\Gamma(\frac{N-1}{2})}{\sqrt{\pi}}\frac{1}{(t^2+x^2)^{\frac{N-1}{2}}}) \\
    & = \frac{\Gamma(\frac{N-1}{2})}{2\pi^{\frac{N+1}{2}}}\frac{1}{(t^2+x^2)^{\frac{N-1}{2}}} \\
    & = \frac{\Gamma(\frac{N-1}{2})}{2\pi^{\frac{N+1}{2}}}\frac{1}{(t^2+|\mat{x}|^2)^{\frac{N-1}{2}}},
\end{aligned}
\end{equation}
which is the Poisson kernel in PFGM~\citep{xu2022poisson}. For general $\mat{x}'$, we have
\begin{equation}
    G(\mat{x},t;\mat{x}) = \frac{\Gamma(\frac{N-1}{2})}{2\pi^{\frac{N+1}{2}}}\frac{1}{(t^2+|\mat{x}-\mat{x}'|^2)^{\frac{N-1}{2}}}
\end{equation}

For general data distribution $p_{\rm data}(\mat{x})$, we have $\phi(\mat{x},t)=\int p_{\rm data}(\mat{x}')G(\mat{x},t;\mat{x}')d^N\mat{x}'$. We now check the three conditions:

{\bf (C1)} holds. 

{\bf (C2)} holds. Although $F(\mat{x}_1',\mat{x}_2',t)$ may not have a closed form for any $t>0$, we notice that for large $T$:
\begin{equation}
    p(\mat{x},T;\mat{x}')\sim \frac{1}{(1+\frac{|\mat{x}-\mat{x}'|^2}{T^2})^{\frac{N+1}{2}}}\approx {\rm exp}(-\frac{(N+1)|\mat{x}-\mat{x}'|^2}{2T^2}),
\end{equation}
so we can show $\underset{T\to\infty}{\rm lim} F(\mat{x}_1',\mat{x}_2',T)\equiv\int\sqrt{p(\mat{x},T;\mat{x}_1')}\sqrt{p(\mat{x},T;\mat{x}_2')}d^N\mat{x}\to 1$ similar to the diffusion equation case.

\subsection{Wave equation}

The Green's function $G(\mat{x},t)$ of the wave equation  for $(\mat{x}\in\mathbb{R}^N)$ satisfies:
\begin{equation}\label{eq:wave}
    G_{tt} - \nabla^2G = \delta(\mat{x})\delta(t),
\end{equation}
where for simplicity we have assumed the source point $\mat{x}'=\mat{0}$. 
Using Fourier transformation Eq.~(\ref{eq:fourier}) , the transformed PDE becomes:
\begin{equation}
     \tilde{G}_{tt} + |\mat{k}|^2 \tilde{G}=0,\quad \tilde{G}_t(\mat{k},0) = 1, \quad \tilde{G}(\mat{k},0) = 0
\end{equation}
whose solution is $\tilde{G}=\frac{1}{|\mat{k}|}({\rm sin}(|\mat{k}|t)$. Now transforming back to $\mat{x}$ space:

\begin{equation}
\begin{aligned}
    G(\mat{x},t) &= \mathcal{F}^{-1}[\tilde{G}] \\ 
    &=\frac{1}{(2\pi)^N}\int_{\mat{k}} \frac{1}{|\mat{k}|}{\rm sin}(|\mat{k}| t){\rm exp}(i\mat{k}\cdot\mat{x})d^N\mat{k} \\
    &=\frac{A_{N-2}\sqrt{\pi}\Gamma(\frac{N-1}{2})}{(2\pi)^N}(\int_{k=0}^\infty k^{N-2}{\rm sin}(k t) {}_0\tilde{\Gamma}_1(\frac{N}{2}, -\frac{(kx)^2}{4})) \\
\end{aligned}
\end{equation}
Doing the integration is not easy, but  
we refer readers to~\cite{soodak1993wakes} for the ideal wave solutions in arbitrary $N$ dimensions, and~\cite{wave_green} for dissipative waves. Here we summarize the main results in ~\cite{soodak1993wakes}. We denote the Green's function in $N$-dimensions by $\phi_n$, then solutions differing by 2 dimensions are related ($r\equiv |\mat{x}|$):
\begin{equation}
    G_{N+2}(r,t) = -\frac{1}{2\pi r}\frac{\partial G_N(r,t)}{\partial r}
\end{equation}
The solutions for $N\leq 5$ are listed ($\tau\equiv t-r$):
\begin{equation}
    \begin{aligned}
        &G_1 = \frac{1}{2}\Theta(\tau) \\
        &G_2 = \frac{1}{2\pi}\frac{\Theta(\tau)}{\sqrt{t^2-r^2}}\\
        &G_3 = \frac{1}{4\pi}\frac{\delta(\tau)}{r} \\
        &G_4 = \frac{1}{4\pi^2}(\frac{\delta(\tau)}{r(t^2-r^2)^{\frac{1}{2}}}-\frac{\Theta(t-r)}{(t^2-r^2)^{\frac{3}{2}}}) \\
        &G_5 = \frac{1}{8\pi^2}(\frac{\delta(\tau)}{r^3}+\frac{\delta'(\tau)}{r^2})
    \end{aligned}
\end{equation}
One interesting observation is that solutions in even and odd dimension have qualitative differences. When $N=1,3,5,...$ is odd, the solution only contains $\delta(\tau)$ and its derivatives (the only exception is $N=1$), meaning only the wave front $t=r$ is excited, with everywhere else zero. By contrast, when $N=2,4,...$ is even, the solution contains the step function $\Theta(\tau)$, which does not vanish for $t>r$, is referred to as ``wake"~\citep{soodak1993wakes}. For general data distribution $p_{\rm data}(\mat{x})$, we have $\phi(\mat{x},t)=\int p_{\rm data}(\mat{x}')G(\mat{x},t;\mat{x}')d^N\mat{x}'$.

{\bf (C1)} fails. In Table \ref{tab:1}, we match $p\equiv -\phi_t$, but is $p$ a valid probability density distribution? We check the 2D case
\begin{equation}
    p \equiv -\phi_t = \frac{1}{2\pi}(\frac{t\Theta(t-r)}{(t^2-r^2)^{\frac{3}{2}}} -\frac{\delta(t-r)}{\sqrt{t^2-r^2}}).
\end{equation}
The second term is negative. Beyond the wave front $r>t$, $p=0$, so $\mat{v}=\frac{\nabla\phi}{p}$ is ill-defined.

{\bf (C2)} fails. For $\mat{x}_1'\neq \mat{x}_2'$, even the support of $p(\mat{x},t;\mat{x}_1')$ and $p(\mat{x},t;\mat{x}_2')$ do not match.

\subsection{Helmholtz equation}

The Helmholtz equation is $(\nabla^2+k_0^2)\phi=0$. It is easy to see that $k_0=0$ recovers the Poisson equation. From the physics perspective, the Helmholtz equation can be interpreted as  single-frequency wave equation, or single-energy Schr\"{o}dinger equation.

{\bf Relation to wave equation} The wave equation is $\phi_{tt}-\nabla^2\phi=0$. It is usually interesting to study periodic solutions (in time) $\phi(\mat{x})=f(\mat{x},t)e^{-ik_0 t}$ with the angular frequency $k_0$. Inserting the ansatz to the wave equation gives $(\nabla^2+k_0^2)f=0$, which is the Helmholtz equation. 

{\bf Relation to Schr\"{o}dinger equation} The Schr\"{o}dinger equation of a free particle is $i\hbar\frac{\partial \phi}{\partial t} = -\frac{\hbar^2}{2m}\nabla^2\phi$ where $\phi$ is the wave function. It is usually interesting to study steady states, such that $\phi(\mat{x},t)=f(\mat{x})e^{-iEt/\hbar}$. Inserting the steady-state ansatz to the Schr\"{o}dinger equation gives $(\nabla^2 + \frac{2mE}{\hbar^2}) f = 0$, which is the Helmholtz equation with $k_0=\frac{\sqrt{2mE}}{\hbar}$.

The Green's function is the solution to (for simplicity, we set $\mat{x}'=\mat{0}$):
\begin{equation}
    \nabla^2G(\mat{x}) + k_0^2G(\mat{x})=\delta(\mat{x})
\end{equation}
The solution can be found at ~\cite{2255877}:
\begin{equation}
    G(\mat{x}) = \frac{i}{4}(\frac{k_0}{2\pi r})^{\frac{N}{2}-1}H_{\frac{N}{2}-1}^{(1)}(k_0r),\quad r\equiv|\mat{x}|
\end{equation}
where $H^{(1)}$ is first kind Hankel function. $H_n^{(1)} \equiv  J_n+iY_n$ where $J_n$ and $Y_n$ are first-kind and second-kind Bessel functions, respectively. When $k\to 0$, $kr\to 0$, $H^{(1)}_{\frac{n}{2}-1}(k_0r)\approx -\frac{i\Gamma(\frac{N}{2}-1)}{\pi}(\frac{2}{k_0r})^{\frac{N}{2}-1}$, so $\phi(\mat{x})\sim \frac{1}{r^{N-2}}$, which is the Green's function of the Poisson equation. Similar to the Poisson equation, we identify one dimension in $\mat{x}$ as time $t$, and we change $N\to N+1$, $\mat{x}\to\mat{x}'=[\mat{x},t]$. This means the equation
\begin{equation}
    G_{tt} + \nabla^2G + k_0^2G = \delta(\mat{x}-\mat{x}'),\quad \mat{x}\in\mathbb{R}^N
\end{equation}
has the solution
\begin{equation}
    G(\mat{x},t;\mat{x}') = \frac{i}{4}(\frac{k_0}{2\pi \sqrt{t^2+r^2}})^{\frac{N-1}{2}}H_{\frac{N-1}{2}}^{(1)}(k_0\sqrt{t^2+r^2}),\quad r\equiv|\mat{x}-\mat{x}'|.
\end{equation}
For simplicity, we only consider the real part (which is still a solution to the Helmholtz equation):
\begin{equation}
    G(\mat{x},t;\mat{x}') = -\frac{1}{4}(\frac{k_0}{2\pi \sqrt{t^2+r^2}})^{\frac{N-1}{2}}Y_{\frac{N-1}{2}}(k_0\sqrt{t^2+r^2}),\quad r\equiv|\mat{x}-\mat{x}'|.
\end{equation}
For general data distribution $p_{\rm data}(\mat{x})$, we have $\phi(\mat{x},t)=\int p_{\rm data}(\mat{x}')G(\mat{x},t;\mat{x}')d^N\mat{x}'$.

{\bf (C1)} Conditionally holds. Note $G$ is a decreasing function of $t$ for small $(t,r)$, because $\frac{k_0}{2\pi\sqrt{t^2+r^2}}$ decreases with increasing $t$, $-Y_{\frac{N-1}{2}}(k_0\sqrt{t^2+r^2})$ is a decreasing (positive) function of $t$ for some range $k_0\sqrt{t^2+r^2}\leq r_c$, where $r_c$ is the first zero of $Y_{\frac{N-1}{2}}$. So $p(\mat{x},t,\mat{x}')\equiv -\phi_t(\mat{x},t;\mat{x}')\geq 0$ for $k_0\sqrt{t^2+r^2}\leq r_c$. When $k_0\sqrt{t^2+r^2} > r_c$, $Y_{\frac{N-1}{2}}(k_0\sqrt{t^2+r^2})$ oscillates and can change sign. For the $(t,r)$ region one is interested in, it suffices to choose $k_0\leq \frac{r_c}{(\sqrt{t^2+r^2})_{\rm max}}$ to make sure that $p$ is a valid density distribution.

{\bf (C2)} Conditionally holds. As long as $k_0$ is small enough (equivalently, phase space is properly clipped), $p$ defines a density distribution.

\subsection{Screened Poisson equation}

The screened Poisson equation is $(\nabla^2-m^2)\phi=0$. The screened Poisson equation is very similar to the Helmholtz equation, the only difference being the sign within the brackets. $m=0$ recovers the Poisson equation. $m$ can be interpreted as the screening magnitude, as in Thomas-Fermi screening or Debye screening; or $m$ can be interpreted as the mass of bosons, as in Yukawa potential.

{\bf Relation to Klein-Gordon equation} The Klein-Gordon (KG) equation is $\phi_{tt}-\nabla^2\phi+m^2\phi=0$. Time-independent solutions (hence $\phi_{tt}=0$) of KG is the screened Poisson equation.

The Green's function is the solution to (for simplicity, we have set $\mat{x}'=\mat{0}$):
\begin{equation}
    \nabla^2G(\mat{x}) - m^2G(\mat{x}) = \delta(\mat{x})
\end{equation}
The solution can be found at~\cite{enwiki:1134903008}:
\begin{equation}
    G(\mat{x}) = -\frac{1}{(2\pi)^{\frac{N}{2}}}(\frac{m}{r})^{\frac{N}{2}-1}K_{\frac{N}{2}-1}(mr),\quad r\equiv |\mat{x}|
\end{equation}
where $K_n$ is the second kind modified Bessel functions. When $x\ll 1$, $K_n(x)\sim \frac{1}{x^n}$. When $m\to 0$, $K_{\frac{N}{2}-1}(mr)\sim \frac{1}{(mr)^{\frac{N}{2}-1}}$, so $G(\mat{x})\sim \frac{1}{r^{N-1}}$ recovers the Poisson kernel.

Similar to the Poisson equation, we identify one dimension in $\mat{x}$ as time $t$, and we change $N\to N+1$, $\mat{x}\to\mat{x}'=[\mat{x},t]$. This means the equation
\begin{equation}
    G_{tt} + \nabla^2G - m^2G = \delta(\mat{x}-\mat{x}'),\quad \mat{x}\in\mathbb{R}^N
\end{equation}
has the solution
\begin{equation}
    G(\mat{x},t;\mat{x}') = \frac{1}{(2\pi)^{\frac{N+1}{2}}}(\frac{m}{\sqrt{t^2+r^2}})^{\frac{N-1}{2}}K_{\frac{N-1}{2}}(m\sqrt{t^2+r^2}),\quad r\equiv |\mat{x}-\mat{x}'|
\end{equation}
For general data distribution $p_{\rm data}(\mat{x})$, we have $\phi(\mat{x},t)=\int p_{\rm data}(\mat{x}')G(\mat{x},t;\mat{x}')d^N\mat{x}'$

{\bf (C1)} holds. $p(\mat{x},t;\mat{x}')\equiv -\phi_t >0$, since both $(\frac{m}{\sqrt{t^2+r^2}})^{\frac{N+1}{2}}$ and $K_{\frac{N-1}{2}}(m\sqrt{t^2+r^2})$ are positive and decreasing functions of $t$.

{\bf (C2)} holds, since the dispersion relation $\omega(k)=-i\sqrt{m^2+k^2}$ satisfies the smoothing condition.

\subsection{Schr\"{o}dinger equation}
The Schr\"{o}dinger equation is $i\phi_t = -\nabla^2 \phi + V(\mat{x}) \phi$, where $\phi$ is the (complex) wave function. It describes the evolution of a quantum particle~\citep{griffiths2018introduction}. For $V(\mat{x})=0$, the free particle Schr\"{o}dinger equation is $i\phi_t = -\nabla^2\phi$. Note that since $\phi$ is a complex scalar function, it is not a valid probability distribution. In fact, quantum mechanics interpret $|\phi|^2$ as the probability distribution. Now we aim to calcuate $\frac{\partial |\phi|^2}{\partial t}$ from the Schr\"{o}dinger equation:
\begin{equation}\label{eq:schrodinger}
    i\phi_t + \nabla^2\phi = 0
\end{equation}
Taking the complex conjugate gives
\begin{equation}\label{eq:schrdinger-cc}
    -i\phi^*_t + \nabla^2\phi^* = 0
\end{equation}
Multiplying $\phi^*$ to Eq.~(\ref{eq:schrodinger}) and multiplying $\phi$ to Eq.~(\ref{eq:schrdinger-cc}) and subtracting the two gives:
\begin{equation}\label{eq:schrodinger-minus}
    i(\phi^*\phi_t + \phi\phi_t^*) + \phi^*\nabla^2\phi - \phi\nabla^2\phi^* = 0.
\end{equation}
Note that 
\begin{equation}
    \phi^*\phi_t + \phi\phi_t^* = \frac{\partial |\phi|^2}{\partial t}, \phi^*\nabla^2\phi - \phi\nabla^2\phi^* = \nabla\cdot (\phi^*\nabla\phi-\phi\nabla\phi^*),
\end{equation}
Eq.~(\ref{eq:schrodinger-minus}) can simplify to
\begin{equation}
    i\frac{\partial |\phi|^2}{\partial t} + \nabla\cdot (\phi^*\nabla\phi-\phi\nabla\phi^*) = i\frac{\partial |\phi|^2}{\partial t} + i\nabla\cdot (|\phi|^2(2{\rm Im}\nabla{\rm log}\phi))=0.
\end{equation}
By comparing to Eq.~(\ref{eq:density-flow}), we have $p=|\phi|^2$, $\mat{v}=2{\rm Im}\nabla{\rm log}\phi$ and $R=0$. The Green's function $G(\mat{x},t;\mat{x}')=\frac{1}{(4it)^{N/2}}{\rm exp}(-\frac{|\mat{x}-\mat{x}'|^2}{4it})$, which can be obtained by simply $t\to it$ in the diffusion kernel. For general data distribution $p_{\rm data}(\mat{x})$, we have $\phi(\mat{x},t)=\int p_{\rm data}(\mat{x}')G(\mat{x},t;\mat{x}')d^N\mat{x}'$.

{\bf (C1)} fails. $p\equiv |\phi|^2$ is a probability distribution, but it may have zeros due to interference, resulting divergence of $\mat{v}$. 

{\bf (C2)} fails. $p(\mat{x},T)$ oscillates restlessly even for large $T$ and dependent on the initial distribution.

\section{Interpolating between DMs and PFGMs}\label{app:interpolating_DM_PFGM}
Let's study this PDE:
\begin{equation}\label{eq:pfgm+dm}
    a\phi_{tt} - b\phi_t + \nabla^2 \phi = \delta(\mat{x})\delta(t),
\end{equation}
where $(a,b)=(1,0)$ and $(a,b)=(0,1)$ recovers the Poisson equation and the diffusion equation, respectively. The Fourier transformation Eq.~(\ref{eq:fourier}) converts the PDE to
\begin{equation}
    a\tilde{\phi}_{tt} - b\tilde{\phi}_t - |\mat{k}|^2\tilde{\phi} = \delta(t),
\end{equation}
or equivalently,
\begin{equation}
    a\tilde{G}_{tt} - b\tilde{G}_t - |\mat{k}|^2\tilde{G} = 0,\quad (a\tilde{G}_t - b\tilde{G})(\mat{k},0) = 1.
\end{equation}
Inserting the trial equation $\tilde{G}= A{\rm exp}(-\omega t)$ to above gives
\begin{equation}
    a\omega^2+b\omega-|\mat{k}|^2 = 0, -\omega a A - bA = 1
\end{equation}
The quadratic equation has the solution $\omega_{\pm}=\frac{1}{2a}(-b\pm\sqrt{b^2+4|\mat{k}|^2})$ where only $w_+$ is consistent with the free boundary condition ($\tilde{\phi}\to 0,$ when $t\to\infty$). So $\omega\equiv \omega_+=\frac{1}{2a}(-b+\sqrt{b^2+4|\mat{k}|^2})$, and $A=-\frac{1}{\omega a+b}=-\frac{2}{b+\sqrt{b^2+4|\mat{k}|^2}}$,
\begin{equation}
    \tilde{G}(\mat{k},t)=-\frac{2}{b+\sqrt{b^2+4|\mat{k}|^2}}{\rm exp}(-\frac{1}{2a}(\sqrt{b^2+4|\mat{k}|^2}-b)t)
\end{equation}
Now we try to transform back to $\mat{x}$ space:

\begin{equation}
\begin{aligned}
    G(\mat{x},t) &= \mathcal{F}^{-1}[\tilde{G}] \\ 
    &=\frac{1}{(2\pi)^N}\int_{\mat{k}} -\frac{2}{b+\sqrt{b^2+4|\mat{k}|^2}}{\rm exp}(-\frac{1}{2a}(\sqrt{b^2+4|\mat{k}|^2}-b)t){\rm exp}(i\mat{k}\cdot\mat{x})d^N\mat{k} \\
    &=-\frac{A_{N-2}}{(2\pi)^N}\int_{k=0}^\infty dk \frac{k^{N-1}}{b+\sqrt{b^2+4k^2}}{\rm exp}(-\frac{1}{2a}(\sqrt{b^2+4k^2}-b) t)\int_{\theta=0}^\pi {\rm exp}(ikx{\rm cos}\theta)({\rm sin}\theta)^{N-2}d\theta \\
    &=-\frac{A_{N-2}\sqrt{\pi}\Gamma(\frac{N-1}{2})}{(2\pi)^N}(\int_{k=0}^\infty \frac{k^{N-1}}{b+\sqrt{b^2+4k^2}}{\rm exp}(-\frac{1}{2a}(\sqrt{b^2+4k^2}-b) t) {}_0\tilde{\Gamma}_1(\frac{N}{2}, -\frac{(kx)^2}{4}))
\end{aligned}
\end{equation}
To the best of the our knowledge and Wolfram Mathematica's ability, the integral does not have a closed from. For general data distribution $p_{\rm data}(\mat{x})$, we have $\phi(\mat{x},t)=\int p_{\rm data}(\mat{x}')G(\mat{x},t;\mat{x}')d^N\mat{x}'$

{\bf (C1)} holds.

{\bf (C2)} holds. The dispersion relation $\omega = \frac{i}{2a}(b-\sqrt{b^2+4ak^2})$ satisfies Eq.~(\ref{eq:C2-dispersion}).

\section{{\bf (C2)} and dispersion relations}\label{app:C2-dispersion}

Define the overlap between $p(\mat{x},t;\mat{x}_i')\ (i=1,2)$:
\begin{equation}\label{eq:F-overlap}
    F(\mat{x}_1',\mat{x}_2',t)\equiv \frac{\int p(\mat{x},t;\mat{x}_1')p(\mat{x},t;\mat{x}_2')d^N\mat{x}}{\int p(\mat{x},t;\mat{x}_1')p(\mat{x},t;\mat{x}_1')d^N\mat{x}},
\end{equation} 
then {\bf (C2)} requires that $\underset{t\to\infty}{\rm lim}F(\mat{x}_1',\mat{x}_2',t)\to 1$. We now attempt to rewrite Eq.~(\ref{eq:F-overlap}) in terms of Fourier bases, where dispersion relation should emerge. Define 
\begin{equation}
    \tilde{p}(\mat{k},t;\mat{x}_i') = \int p(\mat{x},t;\mat{x}_i'){\rm exp}(-i\mat{k}\cdot\mat{x})d^N\mat{x},\quad i=1,2
\end{equation}
Note that $p(\mat{x},t;\mat{x}_i')\ (i=1,2)$ differ only by a translation, so their Fourier function differs only by a phase
\begin{equation}
    p(\mat{k},t;\mat{x}_2') = p(\mat{k},t;\mat{x}_1'){\rm exp}(i\mat{k}\cdot(\mat{x}_1'-\mat{x}_2'))
\end{equation}
Due to the unitarity of Fourier transformation, the integration in $\mat{x}$ can be converted to integration in $\mat{k}$, so
\begin{equation}
\begin{aligned}
    F(\mat{x}_1',\mat{x}_2',t)&=\frac{\int p^*(\mat{k},t;\mat{x}_1')p(\mat{k},t;\mat{x}_2')d^N\mat{k}}{\int p^*(\mat{k},t;\mat{x}_1')p(\mat{k},t;\mat{x}_1')d^N\mat{k}} \\
    & =\frac{\int p^*(\mat{k},t;\mat{x}_1')p(\mat{k},t;\mat{x}_1'){\rm exp}(i\mat{k}\cdot(\mat{x}_1'-\mat{x}_2'))d^N\mat{k}}{\int p^*(\mat{k},t;\mat{x}_1')p(\mat{k},t;\mat{x}_1')d^N\mat{k}} \\
    & = \frac{\int |p(\mat{k},t;\mat{x}_1')|^2{\rm cos}(\mat{k}\cdot(\mat{x}_1'-\mat{x}_2'))d^N\mat{k}}{\int |p(\mat{k},t;\mat{x}_1')|^2 d^N\mat{k}} \\
    & = \langle {\rm cos}(\mat{k}\cdot(\mat{x}_1'-\mat{x}_2'))\rangle_{|p(\mat{k},t;\mat{x}_1')|^2}
\end{aligned}
\end{equation}
where is the expected value of ${\rm cos}(\mat{k}\cdot(\mat{x}_1'-\mat{x}_2'))$ over (unnormalized) distribution $|p(\mat{k},t;\mat{x}_1')|^2$. Note that 
\begin{equation}
    p(\mat{k},t;\mat{x}_1')={\rm exp}(-i\omega(k)t)p(\mat{k},0;\mat{x}_1')={\rm exp}(-i{\rm Re}\ \omega(k)t){\rm exp}({\rm Im}\ \omega(k)t)p(\mat{k},t;\mat{x}_1'),
\end{equation}
we have 
\begin{equation}
    |p(\mat{k},t;\mat{x}_1')|^2 = {\rm exp}(2{\rm Im}\ \omega(k)t)
\end{equation}
where we used $|p(\mat{k},0;\mat{x}_1')|^2=1$. Define
\begin{equation}
    k^*\equiv \underset{k}{\rm argmax}\ {\rm Im}\ \omega(k),
\end{equation}
as $t$ increases, $|p(\mat{k},t;\mat{x}_1')|^2$ has increasingly more probability concentrated around $k=k^*$. As a result,
\begin{equation}
    \underset{t\to\infty}{\rm lim}\ F(\mat{x}_1',\mat{x}_2', t) = \langle{\rm cos}(\mat{k}\cdot (\mat{x}_1'-\mat{x}_2'))\rangle_{|\mat{k}|=k^*},
\end{equation}
where the averaging is over the sphere $|\mat{k}|=k^*$. The limit is 1 if and only if $k^*=0$ for $\mat{x}_1'\neq\mat{x}_2'$. $k^*=0$ is equivalent to
\begin{equation}\label{eq:C2-dispersion-app}
\boxed{
    {\rm Im}\ \omega(k) < {\rm Im}\ \omega(0),\quad {\rm for\ all\ }k>0.
}
\end{equation}
The physical interpretation is that waves of $k>0$ should decay faster than $k=0$.

\section{Examples (long table)}

\begin{table}[tbp]
    \centering
    \begin{adjustbox}{angle=90}
    \resizebox{1.6\textwidth}{!}{
    \begin{tabular}{|c|c|c|c|c|c|c|c|}\hline
     equation  & diffusion equation &  Poisson equation & ideal wave equation & dissipative wave equation & Helmholtz equation &  screened Poisson equation (Yukawa) & Schr\"{o}dinger equation    \\\hline
     PDE $\hat{L}\phi=0$  & $\phi_t-\nabla^2 \phi=0$ &  $\phi_{tt}+\nabla^2\phi=0$ & 
     $\phi_{tt}-\nabla^2\phi=0$ & $\phi_{tt}+ 2\epsilon\phi_{t}-\nabla^2\phi = 0$ & $\phi_{tt}+\nabla^2\phi+ k_0^2\phi=0$ &  $\phi_{tt}+\nabla^2\phi-m^2\phi=0$ & 
     $i\phi_t+\nabla^2\phi=0$ \\\hline
     Rewritten & $\frac{\partial\phi}{\partial t} + \nabla\cdot (\phi(-\nabla{\rm log}\phi))=0$ & $\frac{\partial (-\phi_t)}{\partial t} +  \nabla\cdot((-\phi_t) (\frac{\nabla\phi}{\phi_t}))=0$ & $\frac{\partial (-\phi_t)}{\partial t} + \nabla\cdot((-\phi_t) (-\frac{\nabla\phi}{\phi_t}))=0$ & \scalebox{0.7}{%
    $\frac{\partial(-\phi_t-2\epsilon\phi)}{\partial t} + \nabla\cdot((-\phi_t-2\epsilon\phi) (\frac{\nabla\phi}{\phi_t+2\epsilon\phi}))=0$ } & $\frac{\partial(-\phi_t)}{\partial t}+\nabla\cdot((-\phi_t)(\frac{\nabla\phi}{\phi_t}))-k_0^2\phi=0$ & $\frac{\partial(-\phi_t)}{\partial t}+\nabla\cdot((-\phi_t)(\frac{\nabla\phi}{\phi_t}))+m^2\phi=0$ & $\frac{\partial|\phi|^2}{\partial t}+\nabla\cdot(|\phi|^2(2{\rm Im}\nabla {\rm log}\phi))=0$\\\hline
     $p$ & $\phi$ & $-\phi_t$ & $-\phi_t$ &  $-(\phi_t+2\epsilon\phi)$ & $-\phi_t$ & $-\phi_t$ & $|\phi|^2$ \\\hline
     $\mat{v}$ & $-\nabla{\rm log}\phi$  & $\frac{\nabla\phi}{\phi_t}$  & $-\frac{\nabla\phi}{\phi_t}$  & $\frac{\nabla\phi}{\phi_t+2\epsilon\phi}$  & $\frac{\nabla\phi}{\phi_t}$ & $\frac{\nabla\phi}{\phi_t}$ & $2{\rm Im}\nabla{\rm log}\phi$
    \\\hline
     $R$ & 0 & 0 & 0 & 0 & $k_0^2\phi$ & $-m^2\phi$ & 0 \\\hline
     $G(r,t)$ & $\frac{1}{(4\pi t)^{\frac{N}{2}}}{\rm exp}(-\frac{r^2}{4t})$ & $\frac{1}{(t^2+r^2)^{\frac{N-1}{2}}}$ & $\frac{1}{\sqrt{t^2-r^2}}\Theta(t-r)\ {\rm(2D)}$ & \scalebox{0.7}{$\frac{e^{-\epsilon t}{\rm cosh}(\epsilon\sqrt{t^2-r^2})}{\sqrt{t^2-r^2}}\Theta(t-r)\ {\rm(2D)}$} & $(\frac{k_0}{\sqrt{t^2+r^2}})^{\frac{N-1}{2}}H^{(1)}_{\frac{N-1}{2}}(k_0\sqrt{t^2+r^2})$ & $(\frac{m}{\sqrt{t^2+r^2}})^{\frac{N-1}{2}}K_{\frac{N-1}{2}}(m\sqrt{t^2+r^2})$ & $\frac{1}{(4\pi it)^{\frac{N}{2}}}{\rm exp}(\frac{ir^2}{4t})$  \\\hline
     $\hat{G}(k,t)$ & ${\rm exp}(-k^2 t)$ & ${\rm exp}(-k t)$ & ${\rm exp}(\pm ikt)$ & \makecell{${\rm exp}(-\epsilon t+i\sqrt{k^2-\epsilon^2}t)\ (k>\epsilon)$ \\ ${\rm exp}(-(\epsilon+\sqrt{k^2-\epsilon^2})t)\ (k\leq \epsilon)$}  $\hat{G}(k,t)$ & \makecell{${\rm exp}(-i\sqrt{k_0^2-k^2}t)\ (k\leq k_0)$ \\ ${\rm exp}(-\sqrt{k^2-k_0^2}t)\ (k>k_0)$} & ${\rm exp}(-\sqrt{k^2+m^2}t)$ & ${\rm exp}(ik^2t)$    \\\hline
     {\bf (C1)} & Yes & Yes & No & Conditionally yes & Conditional yes & Yes & No \\\hline
     {\bf (C2)} & Yes & Yes & No & Conditionally yes & Conditional Yes & Yes & No  \\\hline
     \makecell[c]{Illustration \\ $\phi$} & \begin{minipage}{.3\textwidth}
      \includegraphics[width=\linewidth, height=60mm]{./fig/heat.png}
    \end{minipage}
    & 
    \begin{minipage}{.3\textwidth}
      \includegraphics[width=\linewidth, height=60mm]{./fig/poisson.png}
    \end{minipage}
    &
    \begin{minipage}{.3\textwidth}
      \includegraphics[width=\linewidth, height=60mm]{./fig/wave.png}
    \end{minipage}
    & \begin{minipage}{.3\textwidth}
      \includegraphics[width=\linewidth, height=60mm]{./fig/wave_dissipation.png}
    \end{minipage}
    & \begin{minipage}{.3\textwidth}
      \includegraphics[width=\linewidth, height=60mm]{./fig/helmholtz.png}
    \end{minipage}
    & 
    \begin{minipage}{.3\textwidth}
      \includegraphics[width=\linewidth, height=60mm]{./fig/modified_helmholtz.png}
    \end{minipage}
    &
    \begin{minipage}{.3\textwidth}
      \includegraphics[width=\linewidth, height=60mm]{./fig/schrodinger.png}
    \end{minipage}
    \\\hline
     s-generative? & Yes (Diffusion Models) & Yes (Poisson Flow) & No & Conditionally Yes (large $\epsilon$) & Conditionally yes (small $k$) & Yes & No  \\\hline
    \end{tabular}
    }
    \end{adjustbox}
    \caption{Connections between physical PDEs and generative models. $r\equiv|\mat{x}-\mat{x}'|$, where $\mat{x}'$ and $\mat{x}$ are the source point and the field point, respectively.}
    \label{tab:long-table}
\end{table}

\end{document}

%% file: math_commands.tex
\usepackage{amsmath,amsfonts,bm}

\def\eqref#1{equation~\ref{#1}}

\def\1{\bm{1}}

\def\rvx{{\mathbf{x}}}

\DeclareMathAlphabet{\mathsfit}{\encodingdefault}{\sfdefault}{m}{sl}
\SetMathAlphabet{\mathsfit}{bold}{\encodingdefault}{\sfdefault}{bx}{n}













%% file: iclr2023_workshop.bbl
\begin{thebibliography}{35}
\providecommand{\natexlab}[1]{#1}
\providecommand{\url}[1]{\texttt{#1}}
\expandafter\ifx\csname urlstyle\endcsname\relax
  \providecommand{\doi}[1]{doi: #1}\else
  \providecommand{\doi}{doi: \begingroup \urlstyle{rm}\Url}\fi

\bibitem[Aharonov et~al.(2008)Aharonov, Van~Dam, Kempe, Landau, Lloyd, and
  Regev]{aharonov2008adiabatic}
Dorit Aharonov, Wim Van~Dam, Julia Kempe, Zeph Landau, Seth Lloyd, and Oded
  Regev.
\newblock Adiabatic quantum computation is equivalent to standard quantum
  computation.
\newblock \emph{SIAM review}, 50\penalty0 (4):\penalty0 755--787, 2008.

\bibitem[Aleixo \& Capelas~de Oliveira(2008)Aleixo and Capelas~de
  Oliveira]{wave_green}
Rafael Aleixo and Edmundo Capelas~de Oliveira.
\newblock Green's function for the lossy wave equation.
\newblock \emph{Revista Brasileira de Ensino de Física}, 30:\penalty0
  1302.1--1302.5, 01 2008.
\newblock \doi{10.1590/S1806-11172008000100003}.

\bibitem[Courant \& Hilbert(2008)Courant and Hilbert]{courant2008methods}
Richard Courant and David Hilbert.
\newblock \emph{Methods of mathematical physics: partial differential
  equations}.
\newblock John Wiley \& Sons, 2008.

\bibitem[Fatras et~al.(2021)Fatras, S'ejourn'e, Courty, and
  Flamary]{Fatras2021UnbalancedMO}
Kilian Fatras, Thibault S'ejourn'e, Nicolas Courty, and R{\'e}mi Flamary.
\newblock Unbalanced minibatch optimal transport; applications to domain
  adaptation.
\newblock \emph{ArXiv}, abs/2103.03606, 2021.

\bibitem[Griffiths \& Schroeter(2018)Griffiths and
  Schroeter]{griffiths2018introduction}
David~J Griffiths and Darrell~F Schroeter.
\newblock \emph{Introduction to quantum mechanics}.
\newblock Cambridge university press, 2018.

\bibitem[Guenther \& Lee(1996)Guenther and Lee]{guenther1996partial}
Ronald~B Guenther and John~W Lee.
\newblock \emph{Partial differential equations of mathematical physics and
  integral equations}.
\newblock Courier Corporation, 1996.

\bibitem[Ho et~al.(2020)Ho, Jain, and Abbeel]{ho2020denoising}
Jonathan Ho, Ajay Jain, and Pieter Abbeel.
\newblock Denoising diffusion probabilistic models.
\newblock \emph{Advances in Neural Information Processing Systems},
  33:\penalty0 6840--6851, 2020.

\bibitem[Hoogeboom et~al.(2022)Hoogeboom, Satorras, Vignac, and
  Welling]{Hoogeboom2022EquivariantDF}
Emiel Hoogeboom, Victor~Garcia Satorras, Cl'ement Vignac, and Max Welling.
\newblock Equivariant diffusion for molecule generation in 3d.
\newblock \emph{ArXiv}, abs/2203.17003, 2022.

\bibitem[(https://math.stackexchange.com/users/218419/mark viola)()]{2255877}
Mark~Viola (https://math.stackexchange.com/users/218419/mark viola).
\newblock Fundamental solution for helmholtz equation in higher dimensions.
\newblock Mathematics Stack Exchange.
\newblock URL \url{https://math.stackexchange.com/q/2255877}.
\newblock URL:https://math.stackexchange.com/q/2255877 (version: 2017-04-30).

\bibitem[Hyv{\"a}rinen(2007)]{Hyvrinen2007SomeEO}
Aapo Hyv{\"a}rinen.
\newblock Some extensions of score matching.
\newblock \emph{Comput. Stat. Data Anal.}, 51:\penalty0 2499--2512, 2007.

\bibitem[Karras et~al.(2022)Karras, Aittala, Aila, and
  Laine]{karras2022elucidating}
Tero Karras, Miika Aittala, Timo Aila, and Samuli Laine.
\newblock Elucidating the design space of diffusion-based generative models.
\newblock \emph{arXiv preprint arXiv:2206.00364}, 2022.

\bibitem[Kirkwood(2018)]{kirkwood2018mathematical}
James Kirkwood.
\newblock \emph{Mathematical physics with partial differential equations}.
\newblock Academic Press, 2018.

\bibitem[LeCun et~al.(2006)LeCun, Chopra, Hadsell, Ranzato, and
  Huang]{LeCun2006ATO}
Yann LeCun, Sumit Chopra, Raia Hadsell, Aurelio Ranzato, and Fu~Jie Huang.
\newblock A tutorial on energy-based learning.
\newblock 2006.

\bibitem[Lin et~al.(2017)Lin, Tegmark, and Rolnick]{lin2017does}
Henry~W Lin, Max Tegmark, and David Rolnick.
\newblock Why does deep and cheap learning work so well?
\newblock \emph{Journal of Statistical Physics}, 168:\penalty0 1223--1247,
  2017.

\bibitem[Lipman et~al.(2022)Lipman, Chen, Ben-Hamu, Nickel, and
  Le]{lipman2022flow}
Yaron Lipman, Ricky~TQ Chen, Heli Ben-Hamu, Maximilian Nickel, and Matt Le.
\newblock Flow matching for generative modeling.
\newblock \emph{arXiv preprint arXiv:2210.02747}, 2022.

\bibitem[Lu et~al.(2019{\natexlab{a}})Lu, Lu, and Nolen]{Lu2019AcceleratingLS}
Yulong Lu, Jianfeng Lu, and James Nolen.
\newblock Accelerating langevin sampling with birth-death.
\newblock \emph{ArXiv}, abs/1905.09863, 2019{\natexlab{a}}.

\bibitem[Lu et~al.(2019{\natexlab{b}})Lu, Lu, and Nolen]{lu2019accelerating}
Yulong Lu, Jianfeng Lu, and James Nolen.
\newblock Accelerating langevin sampling with birth-death.
\newblock \emph{arXiv preprint arXiv:1905.09863}, 2019{\natexlab{b}}.

\bibitem[Martin et~al.(2016)Martin, Reining, and
  Ceperley]{martin_reining_ceperley_2016}
Richard~M. Martin, Lucia Reining, and David~M. Ceperley.
\newblock \emph{Interacting Electrons: Theory and Computational Approaches}.
\newblock Cambridge University Press, 2016.
\newblock \doi{10.1017/CBO9781139050807}.

\bibitem[Mroueh \& Rigotti(2020)Mroueh and Rigotti]{Mroueh2020UnbalancedSD}
Youssef Mroueh and Mattia Rigotti.
\newblock Unbalanced sobolev descent.
\newblock \emph{ArXiv}, abs/2009.14148, 2020.

\bibitem[Parisi(1981)]{Parisi1981CorrelationFA}
Giorgio Parisi.
\newblock Correlation functions and computer simulations (ii).
\newblock \emph{Nuclear Physics}, 205:\penalty0 337--344, 1981.

\bibitem[Poole et~al.(2022)Poole, Jain, Barron, and
  Mildenhall]{Poole2022DreamFusionTU}
Ben Poole, Ajay Jain, Jonathan~T. Barron, and Ben Mildenhall.
\newblock Dreamfusion: Text-to-3d using 2d diffusion.
\newblock \emph{ArXiv}, abs/2209.14988, 2022.

\bibitem[Rissanen et~al.(2022)Rissanen, Heinonen, and
  Solin]{Rissanen2022GenerativeMW}
Severi Rissanen, Markus Heinonen, and A.~Solin.
\newblock Generative modelling with inverse heat dissipation.
\newblock \emph{ArXiv}, abs/2206.13397, 2022.

\bibitem[Rombach et~al.(2021)Rombach, Blattmann, Lorenz, Esser, and
  Ommer]{Rombach2021HighResolutionIS}
Robin Rombach, A.~Blattmann, Dominik Lorenz, Patrick Esser, and Bj{\"o}rn
  Ommer.
\newblock High-resolution image synthesis with latent diffusion models.
\newblock \emph{2022 IEEE/CVF Conference on Computer Vision and Pattern
  Recognition (CVPR)}, pp.\  10674--10685, 2021.

\bibitem[Saharia et~al.(2022)Saharia, Chan, Saxena, Li, Whang, Denton,
  Ghasemipour, Ayan, Mahdavi, Lopes, Salimans, Ho, Fleet, and
  Norouzi]{Saharia2022PhotorealisticTD}
Chitwan Saharia, William Chan, Saurabh Saxena, Lala Li, Jay Whang, Emily~L.
  Denton, Seyed Kamyar~Seyed Ghasemipour, Burcu~Karagol Ayan, Seyedeh~Sara
  Mahdavi, Raphael~Gontijo Lopes, Tim Salimans, Jonathan Ho, David~J. Fleet,
  and Mohammad Norouzi.
\newblock Photorealistic text-to-image diffusion models with deep language
  understanding.
\newblock \emph{ArXiv}, abs/2205.11487, 2022.

\bibitem[Sohl-Dickstein et~al.(2015)Sohl-Dickstein, Weiss, Maheswaranathan, and
  Ganguli]{sohl2015deep}
Jascha Sohl-Dickstein, Eric Weiss, Niru Maheswaranathan, and Surya Ganguli.
\newblock Deep unsupervised learning using nonequilibrium thermodynamics.
\newblock In \emph{International Conference on Machine Learning}, pp.\
  2256--2265. PMLR, 2015.

\bibitem[Song \& Ermon(2019)Song and Ermon]{Song2019GenerativeMB}
Yang Song and Stefano Ermon.
\newblock Generative modeling by estimating gradients of the data distribution.
\newblock \emph{ArXiv}, abs/1907.05600, 2019.

\bibitem[Song et~al.(2020)Song, Sohl-Dickstein, Kingma, Kumar, Ermon, and
  Poole]{song2020score}
Yang Song, Jascha Sohl-Dickstein, Diederik~P Kingma, Abhishek Kumar, Stefano
  Ermon, and Ben Poole.
\newblock Score-based generative modeling through stochastic differential
  equations.
\newblock \emph{arXiv preprint arXiv:2011.13456}, 2020.

\bibitem[Soodak \& Tiersten(1993)Soodak and Tiersten]{soodak1993wakes}
Harry Soodak and Martin~S Tiersten.
\newblock Wakes and waves in n dimensions.
\newblock \emph{American journal of physics}, 61\penalty0 (5):\penalty0
  395--401, 1993.

\bibitem[Tegmark(1996)]{tegmark1996does}
Max Tegmark.
\newblock Does the universe in fact contain almost no information?
\newblock \emph{arXiv preprint quant-ph/9603008}, 1996.

\bibitem[Vincent(2011)]{Vincent2011ACB}
Pascal Vincent.
\newblock A connection between score matching and denoising autoencoders.
\newblock \emph{Neural Computation}, 23:\penalty0 1661--1674, 2011.

\bibitem[{Wikipedia contributors}(2023)]{enwiki:1134903008}
{Wikipedia contributors}.
\newblock Green's function --- {Wikipedia}{,} the free encyclopedia.
\newblock
  \url{https://en.wikipedia.org/w/index.php?title=Green%27s_function&oldid=1134903008},
  2023.
\newblock [Online; accessed 22-January-2023].

\bibitem[Xu et~al.(2022)Xu, Liu, Tegmark, and Jaakkola]{xu2022poisson}
Yilun Xu, Ziming Liu, Max Tegmark, and Tommi Jaakkola.
\newblock Poisson flow generative models.
\newblock \emph{arXiv preprint arXiv:2209.11178}, 2022.

\bibitem[Xu et~al.(2023)Xu, Liu, Tian, Tong, Tegmark, and
  Jaakkola]{Xu2023PFGMUT}
Yilun Xu, Ziming Liu, Yonglong Tian, Shangyuan Tong, Max Tegmark, and
  T.~Jaakkola.
\newblock Pfgm++: Unlocking the potential of physics-inspired generative
  models.
\newblock \emph{ArXiv}, abs/2302.04265, 2023.

\bibitem[YUKAWA \& SAKATA(1937)YUKAWA and SAKATA]{yukawa1937interaction}
Hideki YUKAWA and Shoichi SAKATA.
\newblock On the interaction of elementary particles ii.
\newblock \emph{Proceedings of the Physico-Mathematical Society of Japan. 3rd
  Series}, 19:\penalty0 1084--1093, 1937.

\bibitem[Zeng et~al.(2022)Zeng, Vahdat, Williams, Gojcic, Litany, Fidler, and
  Kreis]{Zeng2022LIONLP}
Xiaohui Zeng, Arash Vahdat, Francis Williams, Zan Gojcic, Or~Litany, Sanja
  Fidler, and Karsten Kreis.
\newblock Lion: Latent point diffusion models for 3d shape generation.
\newblock \emph{ArXiv}, abs/2210.06978, 2022.

\end{thebibliography}
